\documentclass[12pt]{article}

\usepackage[titletoc,title]{appendix}
\usepackage{authblk}
\usepackage{amssymb}
\usepackage{amsmath}
\usepackage{amsfonts}
\usepackage{graphicx}
\usepackage{icomma}

\usepackage{float} 
\usepackage{marvosym}

\usepackage{wasysym}
\usepackage{multirow}
\usepackage{paralist}
\usepackage{bbm}
\usepackage{chngcntr}
\usepackage{xcolor}
\usepackage{pgf}
\usepackage{enumerate}
\usepackage{enumitem}

\usepackage{tikz}
\usepackage{tikz-qtree}
\usetikzlibrary{shapes,arrows}
\usetikzlibrary{shapes.geometric, arrows,positioning,decorations.pathreplacing}

\usepackage{caption}
\usepackage{subcaption}
\usetikzlibrary{decorations.pathreplacing}
\usepackage{pdfrender}
\usepackage{url}

\usepackage{hyperref}
\hypersetup{
	colorlinks=true,
	citecolor=blue,
	linkcolor=blue,
	filecolor=magenta,      
	urlcolor=cyan,
}

\usepackage[margin=1in]{geometry}

\usepackage[export]{adjustbox}
\usepackage{pifont}

\newcommand{\xmark}{\ding{53}}
\newcommand{\hidden}{\boldsymbol{h}} %hidden state
\newcommand{\pay}{Y} %payment
\newcommand{\indpay}{I} %payment indicator
\newcommand{\predpay}{\hat{Y}}%predicted payment
\newcommand{\predprob}{\hat{p}} %predicted prob
\newcommand{\inp}{\boldsymbol{X}} %input

\allowdisplaybreaks

\begin{document}
	
	\title{Micro-level Reserving for General Insurance Claims using a Long Short-Term Memory Network  }

	\author[1]{Ihsan Chaoubi}
	\author[2]{Camille Besse}
	\author[1]{H\'{e}l\`{e}ne Cossette}
	\author[1,2]{Marie-Pier C\^{o}t\'{e}}
	
	\affil[1]{\'{E}cole d'actuariat, Universit\'{e} Laval, Qu\'{e}bec, Canada}

	\affil[2]{Institut intelligence et donn\'{e}es, Qu\'{e}bec, Canada}

	\maketitle

	\date{}

\begin{abstract}
Detailed information about individual claims are completely ignored when insurance claims data are aggregated and structured in development triangles for loss reserving. In the hope of extracting predictive power from the individual claims characteristics, researchers have recently proposed to move away from these macro-level methods in favor of micro-level loss reserving approaches. We introduce a discrete-time individual reserving framework incorporating granular information in a deep learning approach named Long Short-Term Memory (LSTM) neural network. At each time period, the network has two tasks: first, classifying whether there is a payment or a recovery, and second, predicting the corresponding non-zero amount, if any. We illustrate the estimation procedure on a simulated and a real general insurance dataset. We compare our approach with the chain-ladder aggregate method using the predictive outstanding loss estimates and their actual values. Based on a generalized Pareto model for excess payments over a threshold, we adjust the LSTM reserve prediction to account for extreme payments. 
\end{abstract}


\section{Introduction}

Predicting outstanding liabilities is essential to ensure an insurance company's solvency. The reserving exercise consists in accurately estimating future loss payments that will be made on incurred claims. Traditional macro-level loss reserving methods rely on simple assumptions and run-off triangles with claim informations summarized by occurrence and development periods. This results in a loss of the individual claim characteristics, which may limit the robustness of these models, particularly in situations where portfolio mix or claim characteristics are evolving over time. For an overview of aggregate reserving models, see, e.g., \cite{england2002stochastic} or \cite{wuthrich2008stochastic}. Recently, researchers have proposed various micro-level loss reserving approaches that use detailed individual claims information  in order to improve prediction accuracy. Their main advantage consists in yielding the predicted outstanding loss of each claim based on its individual development and characteristics. Several individual models, in continuous or discrete-time, have been proposed in the literature. Pioneer work by \cite{arjas1989claims}, \cite{norberg1993prediction, norberg1999prediction}, and \cite{haastrup1996claims} set the stage for micro-level reserving. See, e.g., \cite{charpentier2016macro} for an econometric comparison between some aggregate and individual reserving models.

In recent years, individual models have attracted interest, and many contributions achieved promising results. \cite{taylor2008individual} used a discrete-time framework to model individual claim developments, based on claim payments and incurred losses. \cite{zhao2009semiparametric} investigated the occurrence time and the reporting delay for individual claims using semi-parametric models with covariates, then \cite{zhao2010applying} extended this study by incorporating a dependence structure. \cite{pigeon2013individual, pigeon2014individual} proposed a parametric individual reserving model using the multivariate skew-normal distribution for claim payments, in discrete time. In a continuous-time framework, \cite{antonio2014micro} present a semi-parametric individual reserving model including detailed information on the payment rate.  All these granular  models are parametric or semi-parametric, based on fixed structural forms and can potentially be over-parametrized.

Machine learning techniques are highly flexible for handling structured and unstructured data. It is thus no surprise that they are gaining in popularity as building blocks of individual loss reserving models. Many recent contributions showed how machine learning can improve an individual reserving model's prediction accuracy. \cite{wuthrich2018machine} used regression trees to predict the number of payments. Tree-based techniques like ExtraTrees and XGBoost have been applied to predict outstanding individual losses, see, e.g., \cite{baudry2019machine} and \cite{duval2019individual}. 

With the growth of individual claims data collection, storage and improved computing power, it becomes interesting to consider sophisticated forms of machine learning, such as deep neural networks (NNs). The latter require few restrictions and assumptions, incorporate complex non-linear trends, and have high predictive performance. NNs with various architectures have recently been applied to individual claims loss prediction. \cite{wuthrich2018machine} and \cite{taylor2019loss} surveyed recent developments in individual reserving models involving NNs. \cite{gabrielli2020neuralembd} proposed to initialize the NN with a regression model, such as an over-dispersed Poisson, to better align with traditional actuarial practice. \cite{gabrielli2020individual} structured a single NN carrying out simultaneous regression and classification tasks to predict expected payments. Summaries of the claim history are the inputs of his network, whereas  \cite{delong2020neural} used entire claim histories in their NN to predict the joint development of claim occurence and individual payments. Building on this work, \cite{delong2021collective} construct several NNs to estimate reserves for both reported and incurred but not reported claims with smaller architectures, thus reducing the computation time.

\begin{figure}[t]
	\centering
	\begin{footnotesize}
		\begin{tikzpicture}[scale=0.8,node distance=2.3cm]
		\tikzstyle{input} = [rectangle, minimum size= 1.2cm,text centered, draw=blue!40!gray, line width=0.5mm]
		\tikzstyle{output} = [circle, minimum width=1.2cm, minimum height=1cm,text centered, draw=red!80!black,line width=0.5mm]
		\tikzstyle{lstm} = [rectangle, minimum width=1.8cm, minimum height=1.5cm, text centered, draw=green!60!black,line width=0.5mm, rounded corners=2ex]
		\tikzstyle{arrow} = [ultra thick,->,>=stealth]

		%t=1
		\node (lstm1) [lstm, align=center] {RNN \\ module};
		\node (start1) [input, below of=lstm1] {$\inp_1$};
		\node (output1) [output, above of=lstm1] {$\hidden_1$};
		
		\draw [arrow] (start1) -- (lstm1);
		\draw [arrow] (lstm1) -- (output1);
		
		%t=2
		\node (lstm2) [lstm, node distance=3cm,right of=lstm1, align=center] {RNN \\ module};
		\node (start2) [input, below of=lstm2] {$\inp_2$};
		\node (output2) [output, above of=lstm2] {$\hidden_2$};
		
		\draw [arrow] (start2) -- (lstm2);
		\draw [arrow] (lstm2) -- (output2);
		\draw [arrow] (output1.east) -- +(0.8, 0)  
		|- (lstm2);

		%t=n
		\node (lstmn) [lstm, node distance=5cm,right of=lstm2, align=center] {RNN \\ module};
		\node (startn) [input, below of=lstmn] {$\inp_n$};
		\node (outputn) [output, above of=lstmn] {$\hidden_n$};
		
		\node (dots) [node distance=3cm,right of=lstm2] {\huge{$\ldots$}};
		\node (dotsbelow) [ node distance=3cm,right of=start2] {\huge{$\ldots$}};
		
		\draw [arrow] (startn) -- (lstmn);
		\draw [arrow] (lstmn) -- (outputn);
		\draw [arrow] (output2.east) -- +(0.8, 0) |- (dots);
		\draw [arrow] (dots.east) -- (lstmn);
		\path [arrow] (output2.south east) 
		-- +(0.7, 0) |- (dotsbelow.west);

		\end{tikzpicture}
	\end{footnotesize}
	\caption{Structure of an RNN with dynamic input. }\label{LSTM}
\end{figure}
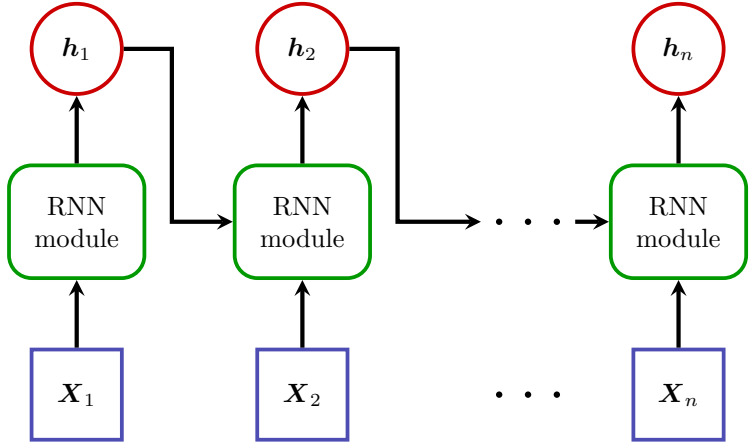

Another way to consider past claim histories is through recurrent neural networks (RNN), a popular class of NNs introduced by \cite{hopfield1982neural}. RNNs are constructed by stacking multiple copies of the same network, allowing information to be passed from one step to the successive network. Figure~\ref{LSTM} shows a typical representation of a RNN, with dynamic input $\inp_t$ and output $\hidden_t$ at period~$t$, for $t=1,\ldots,n$. By construction, RNNs consider relevant features to extract temporal dependencies. However, in basic RNNs, it is challenging to capture long-term dependencies, that are useful to understand claim development, because of the multiplicative gradient that can grow exponentially. To avoid exploding gradient, \cite{hochreiter1997long} introduced Long Short-Term Memory (LSTM) networks, a class of RNNs. The LSTM module has a complex architecture with multiple layers interacting in a special way, and characterized by repeatedly updating memory cells. For several sequential information processing tasks, LSTM networks showed competitive results over simpler RNNs. For more details on LSTMs, see, e.g., \cite{sundermeyer2012lstm}, \cite{graves2013generating} and \cite{weninger2015speech}. In the individual reserving litterature, \cite{kuo2020individual} proposed a multi-period bayesian mixture density network based on LSTMs. Unfortunately, this model does not improve the accuracy of predictions compared to the classic chain-ladder.

In this paper, we develop a new individual loss reserving model applicable for any individual claims dataset with long-tailed development. We focus on the Reported, But Not yet Settled (RBNS) claims by modeling the development of their future payments.  Our model involves an LSTM network which performs two tasks in order to predict expected future payments: a classification to determine the probability of payment or recovery in a given period, and a regression to predict the incremental payment or recovery, given that it is non-zero in a period. A loss-balancing technique allows the simultaneous learning of these two network tasks; the LSTM loss could be adapted to learn more than two tasks and predict several stochastic variables, if required. We implement the model and explore its performance on both a simulated and a real complex dataset. We compare the prediction accuracy with that of a chain-ladder model, similarly as what is done in recent contributions on individual reserving see, e.g., \cite{delong2020neural} and \cite{delong2021collective}. Also, to improve the prediction of large incremental payments, we design a reserving approach that combines the LSTM output and a generalized Pareto distribution. We study its performance on the real dataset. 

Compared to \cite{gabrielli2020individual}, our network is recurrent and uses exact individual claim histories instead of summarized past information. Our model has the advantage of making predictions for the entire future development rather than only yielding an ultimate value. In the case studies presented in Sections~\ref{sec_simulated} and \ref{sec_real_data}, our network outperforms the results obtained with the aggregate chain-ladder, which is an improvement over the model developed in \cite{kuo2020individual}.

The remainder of the paper is organized as follows. In Section~\ref{sec:model}, we introduce  the notation, the architecture and the training process of our individual loss reserving model. Sections~\ref{sec_simulated} and \ref{sec_real_data} present the experiments conducted on a simulated and a real detailed claim dataset, respectively. For these case studies, we explain the model construction, choose the loss function, analyze results and compare them with those of chain-ladder. Also, we describe our reserving approach for large claims in Section~\ref{sec_real_data} and we illustrate it on real data. We conclude in Section~\ref{sec:conclusion} and provide  additional details on the simulated data, real data pre-processing and extreme value model in three Appendices.

\section{Individual loss reserving model}
\label{sec:model}

In Section~\ref{subsec:notation}, we introduce the notation for a generic detailed insurance claims dataset. We then present the architecture and training of our LSTM network in Sections~\ref{subsec:model} and \ref{ss_train}, respectively.

\subsection{Notation}\label{subsec:notation}

\begin{figure}[t]
	\centering
	\begin{tikzpicture}[scale=1]
	
	\draw[thick,decorate,decoration={brace,amplitude=6pt,raise=0pt,mirror}] (1,-0.15) -- (2.9,-0.15);

	\node[align=center] at (1,1.55) {claim\\occurence};
	\draw [thick, ->] (1,1) -- (1,0.1);
	\node[align=center] at (3,2) {reporting};
	\draw [thick, ->] (3,1) -- (3,0.1);
	\node[align=center] at (4,1.55) {opening};
	\draw [thick, ->] (4,1) -- (4,0.1);
	\node[align=center] at (2,-0.9) {reporting \\delay};
	\node[align=center] at (3.7,-0.9) {opening \\lag};
	\node[align=center] at (4.9,1) {\$};
	\draw [thick, ->] (4.9,0.8) -- (4.9,0.1);
	\node[align=center] at (5.5,1) {\$};
	\draw [thick, ->] (5.5,0.8) -- (5.5,0.1);
	\node[align=center] at (6.5,1) {$\ldots$};
	\node[align=center] at (8,1) {\$};
	\draw [thick, ->] (8,0.8) -- (8,0.1);
	
	\draw[thick,decorate,decoration={brace,amplitude=6pt,raise=0pt,mirror}] (4.9,-0.15) -- (8,-0.15);
	
	\draw[thick,decorate,decoration={brace,amplitude=6pt,raise=0pt,mirror}] (3.1,-0.15) -- (4,-0.15);
	
	\node[align=center] at (6.25,-0.9) {payments\\ \& recoveries};
	
	\node[align=center] at (8.5,1.55) {closure};
	\draw [thick, ->] (8.5,1) -- (8.5,0.1);

	\node[align=center] at (11,1.55) {reopening};
	\draw [thick, ->] (11,1) -- (11,0.1);
	
	\node[align=center] at (12,1) {\$};
	\draw [thick, ->] (12,0.8) -- (12,0.1);
	\node[align=center] at (13,1) {\$};
	\draw [thick, ->] (13,0.8) -- (13,0.1);
	
	\node[align=center] at (14,1.55) {final \\ closure};
	\draw [thick, ->] (14,1) -- (14,0.1);
	
	\draw[thick,decorate,decoration={brace,amplitude=6pt,raise=0pt,mirror}] (11,-0.15) -- (14,-0.15);
	\node[align=center] at (12.5,-0.9) {reopening\\period};
	\draw [thick,->] (0,0) -- (15.5,0);
	\node[align=center] at (16,-0.5) {time};

	\draw[thick,decorate,decoration={brace,amplitude=6pt,raise=0pt,mirror}] (1,-1.5) -- (14,-1.5);
	\node[align=center] at (7.5,-2) {claim development};
	\end{tikzpicture}
	\caption{Typical general insurance claim  development.}\label{lifetime}
\end{figure}
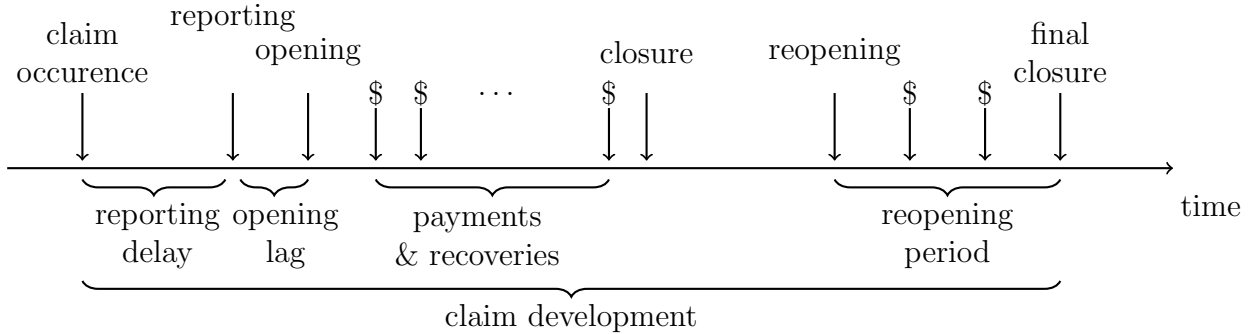

We consider a dataset of $m$ reported general insurance claims. Each claim has a lifetime development; a typical example is depicted in Figure~\ref{lifetime}. A claim occurring within the policy period is usually declared to the insurer with a reporting delay. There may also be an opening lag, which is the delay between the reporting and opening dates. At opening time, specific features about the claim are known, such as the date of occurrence, reporting delay, age of the claimant (body injury insurance), or type of dwelling (home insurance). We denote by $\boldsymbol{S}_{k}$ the set including this static information for claim~$k$, for $k=1, \ldots,m$. Over a claim lifetime, several loss payments and recoveries may be made at different time intervals before closure. A claim can be reopened and then closed again. Throughout this settlement process, we observe additional claim characteristics.

We consider discrete-time claim developments, as the exact daily modeling is too granular and involves weekly variations that are immaterial for reserving. The development periods are of equal length and may be, e.g., months, quarters, or years. We also assume that all claims are fully settled at the end of an ultimate period $n$. For claim $k$, we denote by $\boldsymbol{D}_{k,j}$ the set of dynamic information known at period $j$, for $j=1, \ldots,n$. This may include the incremental payment or recovery $Y_{k,j}$, the claim status (whether the claim is open or not), and the number of claimants related to claim~$k$. 

The reserving problem amounts to estimating the outstanding payments for incurred claims at a given evaluation date~$T^{*}$. 
The claims have a different number of known development periods at~$T^{*}$ depending on their occurrence date. Assume that for claim $k$, we have observed $t_k \leq n$ development periods at $T^*$, as illustrated in Figure~\ref{individual_reserve}. The entire dataset can be represented as 
\begin{equation*}
\{\boldsymbol{S}_k,\ \boldsymbol{D}_{k,j}:j=1,\ldots, t_k; \, k=1, \ldots, m\}.
\end{equation*}
Among the set of dynamic variables, we propose to focus on the prediction of the occurrence and amounts of incremental payments. From now on, a non-zero payment can be positive or negative, as it represents the sum of all cash flows that occured in a development period, including payments, recoveries and subrogations. However, a claim may have a large number of development periods without payment or even an ultimate payment equal to $\$0$. Thus, we define the payment indicator $I_{k,j}=\mathbbm{1}_{\{Y_{k,j} \neq 0\}}$, where $\mathbbm{1}_{\{Y_{k,j} \neq 0\}}=1$ if $Y_{k,j} \neq 0$, and $0$ otherwise. The indicator $I_{k,j}$ will be used in our model to assess the probability of having a non-zero payment. The individual reserve $R_{k}$ is the sum of the incremental payments forecasted for the periods $\{t_{k}+1, \ldots, n\}$, and is expressed as
\begin{equation*}
R_{k}=\sum_{j=t_k+1}^{n}I_{k,j} \times Y_{k,j},
\end{equation*}
which also corresponds to the ultimate loss payment minus the cumulative paid amount at period~$t_k$.

In Section~\ref{subsec:model}, we present our individual reserve approach to estimate $R_k$, for $k=1, \ldots, m$, but we first need to split the full dataset into training, validation, and testing datasets. We use the training dataset, denoted by $\mathcal{T}$, to find the model parameters (weights) that will be confirmed using the validation dataset $\mathcal{V}$. We want to stress that our entire methodology is applicable for a training and validation set containing solely information that is known prior to the evaluation date~$T^*$. Therefore, an insurer could use it as is for setting the reserves at the current evaluation date. After the training process, we use the testing dataset that includes information beyond~$T^*$, to verify the model accuracy. This was possible here as the evaluation date was set in the past.

\subsection{Architecture of the LSTM individual loss reserving model}\label{subsec:model}

\begin{figure}[t]
	\centering
	\begin{tikzpicture}[scale=1]

	\draw [thick, ->] (1,0.8) -- (1,0.1);
	\node[align=center] at (1,1) { $Y_{k,1}$};
	\node[align=center] at (0.95,-.4) {1};
	\draw [thick, ->] (2,0.8) -- (2,0.1);
	\node[align=center] at (2,1) {$Y_{k,2}$};
	\node[align=center] at (4.5,1) {\huge{$\ldots$}};
	\node[align=center] at (7,1) {$Y_{k,t_{k-1}}$};
	\draw [thick, ->] (7,0.8) -- (7,0.1);
	\node[align=center] at (8,1) {$Y_{k,t_k}$};
	\draw [thick, ->] (8,0.8) -- (8,0.1);
	\node[align=center] at (8,-.4) {$t_k$};
	
	\draw[thick,decorate,decoration={brace,amplitude=6pt,raise=0pt,mirror}] (1,-0.6) -- (8.3,-0.6);
	\node[align=center] at (5,-1) {payments observed before $T^*$};
	
	\node[align=center] at (8.85,2.8) {$T^{*}$};
	\node[align=center] at (8.5,3.2) {evaluation date};
	\draw [thick, dashed] (8.5,-1) -- (8.5,3);

	\draw [thick, ->, dashed, blue] (9,0.8) -- (9,0.1);
	\node[align=center,blue] at (10,1.5) {$\hat{p}_{k,t_{k}+1} \times \hat{Y}_{k,t_{k}+1}$};
	\node[align=center] at (9.05,-0.4) {$t_{k}+1$};
	
	\node[align=center,blue] at (12.5,1.4) {\huge{$\ldots$}};
	
	\draw [thick, ->, dashed, blue] (13,0.8) -- (13,0.1);
	\node[align=center ] at (14,-0.4) {n};
	\draw [thick, ->, dashed ,blue] (14,0.8) -- (14,0.1);
	\node[align=center,blue] at (14.2,1.5) {$\hat{p}_{k,n} \times \hat{Y}_{k,n}$};
	\draw[thick,decorate,decoration={brace,amplitude=6pt,raise=0pt,mirror}] (8.7,-0.6) -- (14,-0.6);
	\node[align=center] at (11.5,-1) {future payments };

	\draw [thick,->] (0,0) -- (15.5,0);
	\node[align=center] at (16,-0.5) {time};
	\end{tikzpicture}
	\caption{Discrete-time claim development with observed and future payments at~$T^*$.}\label{individual_reserve}
\end{figure}
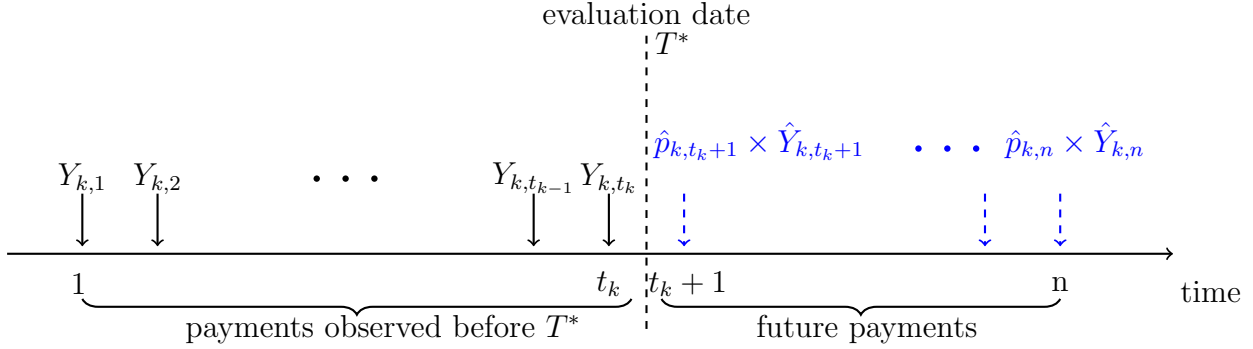

For claim $k \in\{1, \ldots, m\}$, let $\hat{p}_{k,j}$ be the predicted probability of non-zero payment or recovery in period $j\in\{1, \ldots, n\}$, and $\hat{Y}_{k,j}$ the corresponding predicted amount given that it is non-zero. We define the predicted expected incremental payment for claim $k$ in period $j$ by the product $\hat{p}_{k,j} \times \hat{Y}_{k,j}$. At the evaluation date $T^*$, the insurer must predict future payments illustrated in blue in  Figure~\ref{individual_reserve}. We propose an individual reserving approach to estimate $R_k$ with
\begin{equation*}
\hat{R}_{k}=\sum_{j=t_k+1}^{n}\hat{p}_{k,j} \times \hat{Y}_{k,j}.
\end{equation*}

The reserving problem thus amounts to the prediction of the pairs  $\{(\hat{p}_{k,j},\hat{Y}_{k,j}):j=t_k+1, \ldots, n\}$. To achieve this goal, we introduce the LSTM network depicted in Figure~\ref{LSTM_model}, where we drop the claim index $k$ and keep only the time index $j$ for simplicity. Since we study RBNS claims, at least one development period is known for each claim, so the LSTM network contains $n-1$ modules. At step~$j$, we provide the network with the input vector $\boldsymbol{X}_{k,j}=\{\boldsymbol{C}_{k,0},\ \boldsymbol{D}_{k,j}\}$ which is the concatenation of $\boldsymbol{D}_{k,j}$, the dynamic information at period~$j$, and the static context $\boldsymbol{C}_{k,0}$. In the following, we first explain the construction of $\boldsymbol{C}_{k,0}$  and the payment pre-processing within $\boldsymbol{D}_{k,j}$ to form $\boldsymbol{X}_{k,j}$. Then, we describe how inputs are fed to each module as in Figure~\ref{LSTM_model} and we explain the resulting outputs.

\tikzset{sigmoid/.style={path picture= {
			\begin{scope}[x=1pt,y=10pt]
				\draw plot[domain=-7:7] (\x,{1/(1 + exp(-\x))-0.5});
			\end{scope}
		}
	}
}
\begin{figure}[t]
	\centering
	\begin{footnotesize}
		\begin{tikzpicture}[node distance=1.5cm]
		\tikzstyle{replace} = [ minimum size= .4cm, line width=0.8mm, draw=orange, thick, ->,dashed,inner sep=3mm]
		\tikzstyle{input} = [rectangle, minimum size= .7cm,text centered, draw=blue!40!gray, line width=0.5mm]
		\tikzstyle{input2} = [rectangle, dashed, minimum size= .7cm,text centered, draw=orange, line width=0.5mm]
		\tikzstyle{output} = [circle, minimum width=.7cm, minimum height=.7cm,text centered, draw=red!80!black,line width=0.5mm]
		\tikzstyle{lstm} = [rectangle, minimum width=1.8cm, minimum height=1.5cm, text centered, draw=green!60!black,line width=0.5mm,rounded corners=2ex ]
		\tikzstyle{FC} = [trapezium, minimum width=1cm, minimum height=0.05cm, text centered, fill=cyan!50,draw= black,line width=0.1mm,node distance=.3cm, font=\footnotesize ]
		\tikzstyle{sig} = [circle, minimum size= .5cm, text centered,fill=purple!40,draw= black,line width=0.1mm,node distance=0.2cm, sigmoid]
		\tikzstyle{res} = [diamond, minimum size= 1.2cm, text centered, draw=red!50!gray,line width=0.5mm,node distance=0.3cm ]
		\tikzstyle{arrow} = [ultra thick,->,>=stealth]
		\tikzstyle{box}=[draw, minimum width=4.5cm, minimum height=0.1cm]
		
		%t=1
		\node (lstm1) [lstm, align=center] {LSTM \\ module};
		\node (start1) [input, below of=lstm1] {$\inp_1$};
		\node (output1) [output, above of=lstm1] {$\hidden_1$};
		\node (FC11) [FC, above right=of output1]{FC};
		\node (sig1) [sig, above=of FC11] {};
		\node (FC12) [FC, above left=of output1] {FC};
		\node (p1) [res, above=of sig1] {$\predprob_2$};
		\node (y1) [res, above =of FC12,yshift=.71cm] {$\predpay_2^*$};
		%t=2
		\node (lstm2) [lstm, node distance=6cm,right of=lstm1, align=center] {LSTM  \\ module};
		\node (start2) [input, below of=lstm2] {$\inp_2$};
		\node (output2) [output, above of=lstm2] {$\hidden_2$};
		\node (FC21) [FC, above right=of output2]{FC};
		\node (sig2) [sig, above=of FC21] {};
		\node (FC22) [FC, above left=of output2] {FC};
		\node (p2) [res, above=of sig2] {$\predprob_3$};
		\node (y2) [res, above =of FC22,yshift=.71cm] {$\predpay_3^*$};
		\node (start2t) [input2,  align=left,left of=start2,node distance=1.75cm] {$\indpay_2\leftarrow\predprob_2$ \\ $\pay_2^*\leftarrow\predprob_2\times \predpay_2^*$};
		
		%t=i
		\node (lstmi) [node distance=3.8cm,right of=lstm2, align=center] {\huge{$\ldots$}};
		
		%t=n
		\node (lstmn) [lstm, node distance=8cm,right of=lstm2, align=center] {LSTM \\ module};
		\node (startn) [input, below of=lstmn] {$\inp_{n-1}$};
		\node (outputn) [output, above of=lstmn] {$\hidden_{n}$};
		\node (FCn1) [FC,above right=of outputn]{FC};
		\node (sign) [sig, above=of FCn1] {};
		\node (FCn2) [FC,above left=of outputn] {FC};
		\node (pn) [res, above=of sign] {$\predprob_{n}$};
		\node (yn) [res, above =of FCn2,yshift=.71cm] {$\predpay_{n}^*$};
		\node (startnt) [input2,  align=left,left of=startn,node distance=2.35cm] {$\indpay_{n-1}\leftarrow \predprob_{n-1}$ \\ $\pay_{n-1}^*\leftarrow\predprob_{n-1}\times \predpay_{n-1}^*$};

		\draw [arrow] (start1) -- (lstm1);
		\draw [arrow] (lstm1) -- (output1);
		\draw [ultra thick,-] (output1) -- (FC11);
		\draw [ultra thick,-] (output1) -- (FC12);
		\draw [ultra thick,-] (FC11) -- (sig1);
		\draw [arrow] (sig1) -- (p1);
		\draw [arrow] (FC12) -- (y1);
		\draw [orange, thick, dashed,inner sep=2mm]
		([shift={( 1em, 1.9ex)}]p1.north east)
		-- ([shift={(-1em, 1.9ex)}]y1.north west)
		|- ([shift={(-1em, 1.9ex)}]y1.south west)
		|- ([shift={( 1em,-1.9ex)}]p1.south east)
		-- cycle;
		\draw [orange, thick, ->,dashed,inner sep=2mm] (p1.east) -- +(1,0) -- +(1,-5.55) -- ([shift={(-2.5cm,0)}]start2.west);

		\draw [arrow] (lstm1) -- (lstm2);
		\draw [arrow] (start2) -- (lstm2);
		\draw [arrow] (lstm2) -- (output2);
		\draw [ultra thick,-] (output2) -- (FC21);
		\draw [ultra thick,-] (output2) -- (FC22);
		\draw [ultra thick,-] (FC21) -- (sig2);
		\draw [arrow] (sig2) -- (p2);
		\draw [arrow] (FC22) -- (y2);

		\draw [orange, thick, dashed,inner sep=2mm]
		([shift={( 1em, 1.9ex)}]p2.north east)
		-- ([shift={(-1em, 1.9ex)}]y2.north west)
		|- ([shift={(-1em, 1.9ex)}]y2.south west)
		|- ([shift={( 1em,-1.9ex)}]p2.south east)
		-- cycle;

		\draw [orange, thick, dashed,inner sep=2mm]
		([shift={( 1em, 1.9ex)}]pn.north east)
		-- ([shift={(-1em, 1.9ex)}]yn.north west)
		|- ([shift={(-1em, 1.9ex)}]yn.south west)
		|- ([shift={( 1em,-1.9ex)}]pn.south east)
		-- cycle;
		
		\node (dotsbelow) [ node distance=3.6cm,right of=start2] {\huge{$\ldots$}};
	
		\draw [orange, thick, ->,dashed,inner sep=2mm] (p2.east) -- +(1,0) -- +(1,-5.55) -- ([shift={(-4.5cm,0)}]startn.west);
		
		\draw [ultra thick,-] (outputn) -- (FCn1);
		\draw [ultra thick,-] (outputn) -- (FCn2);
		\draw [ultra thick,-] (FCn1) -- (sign);
		\draw [arrow] (sign) -- (pn);
		\draw [arrow] (FCn2) -- (yn);
		\draw [arrow] (startn) -- (lstmn);
		\draw [arrow] (lstmn) -- (outputn);
		\draw [arrow] (lstm2) -- (lstmi);
		\draw [arrow] (lstmi) -- (lstmn);
		\draw [orange, thick, ->,dashed,inner sep=2mm] (start2t) -- (start2);
		\draw [orange, thick, ->,dashed,inner sep=2mm] (startnt) -- (startn);
		
		%legend
		
		\matrix [node distance= 7.5cm, above of=start2,xshift=3cm] {
			\node[input, align=right, label = right: $ \ \ \{ \boldsymbol{C_{0}};\boldsymbol{D}_j\}$]{$ \boldsymbol{X}_j$};\\
			\node[] { }; \\
			\node[replace,label=right: \quad Replacement process ] { }; \\
		};

		\matrix [node distance= 7.5cm, above of=startn,xshift=-0.5cm] {
			\node [FC,align=left,label=right: \ Fully connected] {FC}; \\
			\node[] { }; \\
			\node [sig,label=right: \quad Sigmoid] {}; \\
		};
	
		%			\node[] { }; \\
	\draw[draw=black] (6.7,5.1) rectangle ++(8.7,1.8);
	
		\end{tikzpicture}
	\end{footnotesize}
	\caption{Architecture of the LSTM individual loss reserving network.}\label{LSTM_model}
\end{figure}
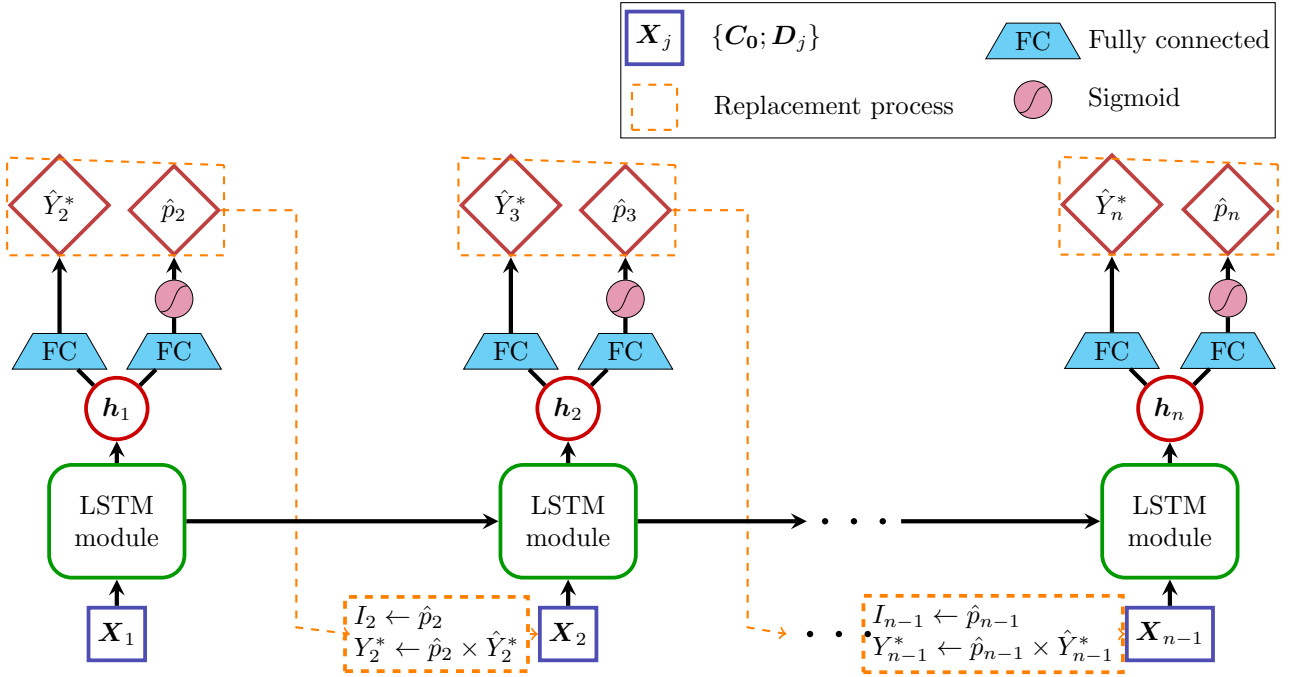

\tikzset{
	neuron/.style={shape=circle,draw,inner sep= 0pt,minimum size = 1.3 em, node 
		distance = 10ex and 1 em,fill=white},
	every loop/.style={min distance=10mm,looseness=5},
	dot/.style={shape=circle,minimum size=1mm, inner sep=0pt, fill=black, node 
		distance= 1ex and 2 em
	},
	group/.style={rectangle,draw, inner sep=1pt,rounded corners,minimum height= 
		3.5em,minimum width=15.5 em, node distance= 1ex and 1em},
	conn/.style={draw,-latex'}
	
}

\begin{figure}[h!]
	\centering
	\begin{footnotesize}
		\begin{tikzpicture}[node distance=2.4cm]
		\tikzstyle{output} = [rounded corners=2ex,rectangle, minimum width=.7cm, minimum height=4cm,text centered, draw=red!80!black,line width=0.2mm]
		\tikzstyle{vect} = [rectangle, minimum width=3cm, minimum height=0.6cm, text centered, draw=blue!60!black,line width=0.2mm,rounded corners=2ex, font=\scriptsize ]
		\tikzstyle{cat} = [ text centered, line width=0.2mm, font=\scriptsize ]
		\tikzstyle{FC} = [rounded corners=2ex,rectangle, minimum width=1cm, minimum height=0.05cm, text centered, fill=cyan!50,draw= black,line width=0.1mm,node distance=.3cm ]
		\tikzstyle{arrow} = [thick,->,>=stealth]
		\tikzstyle{box}=[draw, minimum width=4.5cm, minimum height=0.1cm]

		\node (cat1) [cat, align=center] {Categorical\\feature $\boldsymbol{S}_{k,1}$};
		
		\node [node distance=1.1cm, below of = cat1] {$\huge{\vdots}$};
		\node (cati) [cat, align=center, below of = cat1] {Categorical\\feature $\boldsymbol{S}_{k,i}$};
		\node [node distance=1.1cm, below of = cati] {$\huge{\vdots}$};
		\node (catp) [cat, align=center, below of = cati] {Categorical\\feature $\boldsymbol{S}_{k,p}$};
		
		\node (embd1) [vect, align=center, node distance=3.5cm,right of=cat1 ] {};
		\node [neuron, node distance = 2.6 cm, right of= cat1]{};
		\node [neuron, node distance = 3.2 cm, right of= cat1]{};
		\node [neuron, node distance = 3.8 cm, right of= cat1]{};
		\node [neuron, node distance = 4.4 cm, right of= cat1]{};
		\node[node distance=1.7cm,yshift=-.45cm, right of=embd1]{dim $\ell_1$};

		\node [node distance=1.1cm, below of = embd1] {$\huge{\vdots}$};
		\node (embdi) [vect, align=center, node distance=3.5cm,right of=cati ] {};

		\node [neuron, node distance = 2.6 cm, right of= cati]{};
		\node [neuron, node distance = 3.2 cm, right of= cati]{};
		\node [neuron, node distance = 3.8 cm, right of= cati]{};
		\node [neuron, node distance = 4.4 cm, right of= cati]{};
		\node[node distance=1.7cm,yshift=-.45cm, right of=embdi]{dim $\ell_i$};
		
		\node [node distance=1.1cm, below of = embdi] {$\huge{\vdots}$};
		\node (embdp) [vect, align=center, node distance=3.5cm,right of=catp ] {};

		\node [neuron, node distance = 2.6 cm, right of= catp]{};
		\node [neuron, node distance = 3.2 cm, right of= catp]{};
		\node [neuron, node distance = 3.8 cm, right of= catp]{};
		\node [neuron, node distance = 4.4 cm, right of= catp]{};
		\node[node distance=1.7cm,yshift=-.45cm, right of=embdp]{dim $\ell_p$};
		
		 \node [align=center, above of = embd1, node distance = 1cm] {Embedding vector};

	    \node (concat) [node distance=3cm,right of = embdi] {\Large{$\boldsymbol\otimes$}};
	    
	    \node (stat) [vect, node distance = 3.7cm, align=center, below of = concat] {Scaled quantitative\\features };
	    \node[node distance=1.8cm,yshift=-.45cm, right of=stat]{dim $\ell_q$};
	    	
	    \node (vec_concat) [rounded corners=2ex,rectangle, minimum width=.7cm, minimum height=4cm,text centered, draw=blue!80!black,line width=0.2mm, align=center, node distance = 2cm, right of = concat] {};
	    \node [align=center, node distance = 2.5cm, above of = vec_concat] {Concatenated vector};
	    \node [neuron, node distance = 2 cm, right of= concat]{};
	    	
	    \node [neuron, node distance = 2 cm, right of= concat,yshift=-.6cm]{};
	    \node [neuron, node distance = 2 cm, right of= concat,yshift=.6cm]{};
	    \node [neuron, node distance = 2 cm, right of= concat,yshift=-1.2cm]{};
	    \node [neuron, node distance = 2 cm, right of= concat,yshift=1.2cm]{};
	    \node [node distance= 1.6cm, right of= vec_concat,yshift=-1.6cm]{dim $\sum_{i=1}^{p} \ell_i+\ell_q$};

	    \node (FC) [FC,node distance=4cm,right of = concat] {FC};
			
		\node (output) [output, align=center, node distance = 2cm, right of = FC] {};
		\node [align=center, node distance = 2.5cm, above of = output] {Claim context \\ $\boldsymbol{C}_{k,0}$};
		\node [neuron, node distance = 2 cm, right of= FC]{};
		
		\node [neuron, node distance = 2 cm, right of= FC,yshift=-.6cm]{};
		\node [neuron, node distance = 2 cm, right of= FC,yshift=.6cm]{};
		\node [neuron, node distance = 2 cm, right of= FC,yshift=-1.2cm]{};
		\node [neuron, node distance = 2 cm, right of= FC,yshift=1.2cm]{};
		\node [node distance= 1cm, right of= output,yshift=-1.6cm]{dim $c$};
		
		\draw [arrow] (cat1) -- (embd1);
        \draw [arrow] (cati) -- (embdi);
		\draw [arrow] (catp) -- (embdp);
		\draw [arrow] (embd1) -- (concat);
		\draw [arrow] (embdi) -- (concat);
		\draw [arrow] (embdp) -- (concat);
		\draw [arrow] (stat) -- (concat);
		\draw [arrow] (concat) -- (vec_concat);
		\draw [arrow] (vec_concat) -- (FC);
		\draw [arrow] (FC) -- (output);

		%legend
		\matrix [draw,node distance= 6cm, right of=stat] {
		\node[label=right:\ Concatenation operator] {\large{$\boldsymbol\otimes$} };\\
		\node [FC,align=left,label=right:Fully connected] {FC}; \\
	};
		\end{tikzpicture}
	\end{footnotesize}
	\caption{Static variable engineering for the claim context vector.}\label{embd_figure}
\end{figure}
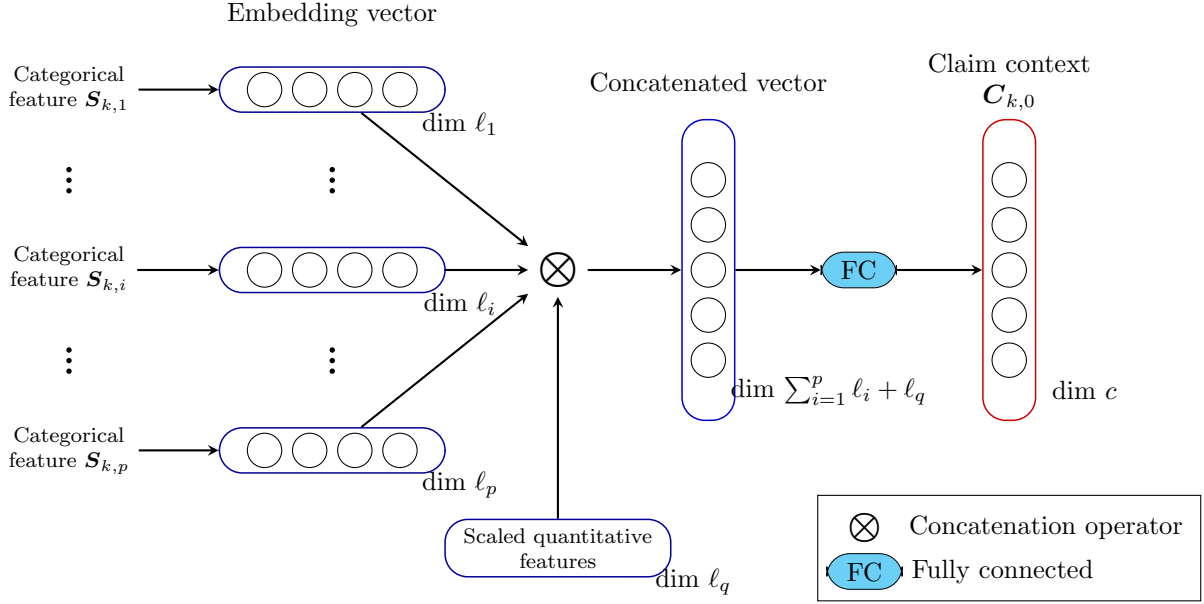

The static features known at the opening date provide a context for the claim and may inform us on its development. The set $\boldsymbol{S}_k$ may include quantitative and/or categorical variables. Each of these types of variables must be prepared differently. Figure~\ref{embd_figure} illustrates the static feature engineering used in this paper. The $q$ quantitative variables are scaled and each of the $p$ categorical variables is indexed using a defined dictionary, then passed to an embedding layer. The latter maps each level of categorical variable $\boldsymbol{S}_{k,i}$ to an $\ell_{i}$-dimensional vector, where $\ell_{i}>1$ is a hyper-parameter that depends on the number of modalities and $i=1,\ldots,p$ (see, e.g., \cite{yin2018dimensionality}). After transformation, the static variables are concatenated and then passed to a linear layer for dimension reduction. We obtain the $k$th claim's context $C_{k,0}$, a vector of dimension $c$, where $c>1$ is a hyper-parameter. Note that we choose to recall the context at each step of the LSTM in Figure~\ref{LSTM_model} to strengthen the network, as the latter will keep only  the relevant information through its particular internal memory cells. 

The set of dynamic features $\boldsymbol{D}_{k,j}$ is composed of the occurrence and amounts of incremental payments, as well as two dynamic variables, namely the development period $j$ and the indicator of whether the period $j$ is observed $r_{k,j}=\mathbbm{1}_{\{j \leq t_k\}}$. Recall that all the LSTM modules have an identical construction with exactly the same weights, so these two deterministic inputs will provide the network with information on the development period being considered. The incremental payments are centered and scaled for the training, validation, and testing datasets as follows 
\begin{equation}\label{transform}
Y_{k,j}^{*}= \left(Y_{k,j}-\mu_{\mathcal{T}}\right)/\sigma_{\mathcal{T}},
\end{equation}
with mean $\mu_{\mathcal{T}}$ and scale $\sigma_{\mathcal{T}}$ computed from all payments (positive and negative) in the training dataset. For our model, we set~$\boldsymbol{D}_{k,j}=\{ j,\ r_{k,j}, \ I_{k,j},\ Y_{k,j}^{*}\}$, for $j=1, \ldots, t_k$. Note that one could include the claim status as a dynamic input, but that means it has to be an extra task for the network. To reduce the model complexity, we ignored this information. In our data, a claim may re-open and close again within a developement period only to allow a payment, so the status closed at the end of two successive periods does not imply that there is no payment in between.

Now that we have presented the construction of the inputs $\boldsymbol{X}_{k,j}$, for $j=1, \ldots, t_k$, we recall Figure~\ref{LSTM_model} to explain the network flow. At period $j$, the vector $\boldsymbol{X}_{k,j}$ passes through the LSTM module whose specific architecture (see, e.g., \cite{yu2019review}) leads to the hidden state vector $\hidden_{k,j}$. We deduce from it the pair of predictions $(\hat{p}_{k,j+1},\hat{Y}_{k,j+1}^{*})$ for the next period $j+1$ as follows. The vector $\hidden_{k,j}$ is passed, in parallel, to a linear layer yielding the payment~$\hat{Y}_{k,j+1}^{*}$, and to another linear layer, then a sigmoid neuron, to obtain the predicted probability~$\hat{p}_{k,j+1}$. Mathematically, we have 
\begin{equation*}
\hat{Y}_{k,j+1}^{*} = \boldsymbol{\beta}  \hidden_{k,j}^{\top}+a, \qquad  \hat{Y}_{k,j+1}=\mu_{\mathcal{T}}+\sigma_{\mathcal{T}}\hat{Y}_{k,j+1}^{*} \qquad \text{  and } \qquad \hat{p}_{k,j+1} = [1+\exp\{-( \boldsymbol{\beta}^\prime  \hidden_{k,j}^{\top}+a^\prime)\}]^{-1},
\end{equation*}
with $\hidden_{k,j}^{\top}$ the transpose of $\hidden_{k,j}$, $\boldsymbol{\beta}$ and $\boldsymbol{\beta}^{\prime}$ two vectors of learnable weights, and $a$ and $a^\prime$ are constants associated to the two linear layers. Here, it is appropriate to use a linear layer for the regression task as we predict the (scaled) payment or recovery, so we allow for negative payments. The entire network's output is the set $\lbrace(\hat{p}_{k,j}, \hat{Y}_{k,j}^{*}), \ j=2, \ldots, n\rbrace$, from which we can retrieve the predictions for the future periods $\lbrace(\hat{p}_{k,j}, \hat{Y}_{k,j}), j=t_k+1, \ldots, n\rbrace$. Moreover, for time step $j$, the target is the pair $(I_{k,j}, Y_{k,j}^{*})$ only if it is observed before the evaluation date. Otherwise, it is an empty pair. We will explain in Section~\ref{ss_train} how the network's optimization is carried out with missing targets. 

An important point is how the network deals with the missing values in the inputs to ensure information propagation. At the first step, the input $\boldsymbol{X}_{k,1}$ is observed and provided to the first LSTM module. From $\hidden_{k,1}$, we obtain the predicted pair  $(\hat{p}_{k,2}, \hat{Y}_{k,2}^{*})$ for the second development period. Moving to the second step, if $t_k<2$, then $I_{k,2}$ and $Y_{k,2}^{*}$ are missing values in the input $\boldsymbol{X}_{k,2}$. They are imputed by the predicted expected values, $\hat{p}_{k,2}$ and $\hat{p}_{k,2}\times \hat{Y}_{k,2}^{*}$, respectively. Otherwise, when $t_k\geq 2$, we keep $\boldsymbol{X}_{k,2}$ as actual observed values. We proceed in the same manner for the other periods until time $n-1$. In Figure~\ref{LSTM_model}, orange dotted lines indicate this replacement process for the inputs. Thus, our network, which is of fixed length, may process sequences with different lengths.

\subsection{Model training}\label{ss_train}

In this section, we describe the training process, the network's two tasks, and the balanced loss function. Also, we explain the regularization techniques and metrics used.

The learning process consists in determining the optimal network weights that minimize an appropriate loss function. We update weights with a stochastic gradient descent at each epoch, relying on all mini-batches obtained by randomly splitting the training dataset. The mini-batch size is a hyper-parameter denoted $b$, and an epoch refers to a complete iteration through all mini-batches. At the end of each epoch, we evaluate the loss function on the validation dataset to measure the network generalization capabilities. Furthermore, given that we produce at each time period the prediction of two variables, our network performs two tasks simultaneously.  

The first task is a binary classification with target $I_{k,j}$, indicating whether the claim~$k$ has a non-zero payment in period~$j$, for $k=1, \ldots, m$ and $j=2, \ldots, n$. The indicator takes value in $\{0,1\}$, so we choose the binary cross-entropy as loss function. For each mini-batch, the loss is defined as the weighted sum of cross-entropy losses per development period. For a given period, the loss calculation involves only observations with non-missing target $I_{k,j}$. As we move forward through the development periods, we get fewer observed data, and hence we need to weigh the sum by the number of available observations. The cross-entropy loss for a mini-batch can be expressed as 
\begin{align}\label{CE}
	CE\left(\boldsymbol{\hat{p}}\right) &= \sum_{j=2}^{n}  \frac{\sum_{i=1}^{b}\delta_{i,j}}{\sum_{h=2}^{n} \sum_{i=1}^{b} \delta_{i,h}} \left(\sum_{k=1}^{b} \frac{\delta_{k,j}}{\sum_{i=1}^{b}\delta_{i,j}} \left[-\left\{\mathbbm{1}_{\{I_{k,j} =0\}} \log\left(1-\hat{p}_{k,j}\right)+\mathbbm{1}_{\{I_{k,j} =1\}} \log\left(\hat{p}_{k,j}\right)\right\}\right]\right) \notag \\
	&= \frac{1}{\sum_{h=2}^{n} \sum_{i=1}^{b} \delta_{i,h}} \sum_{j=2}^{n}   \left(\sum_{k=1}^{b} \delta_{k,j} \left[-\left\{\mathbbm{1}_{\{I_{k,j} =0\}} \log\left(1-\hat{p}_{k,j}\right)+\mathbbm{1}_{\{I_{k,j} =1\}} \log\left(\hat{p}_{k,j}\right)\right\}\right]\right),
\end{align}
with $\boldsymbol{\hat{p}}=\{\hat{p}_{k,j}:k=1, \ldots, b, j=2, \ldots, n\}$ and to only use observations with known target, we set $$\delta_{k,j} = \left\{
\begin{array}{rl}
1, & \text{ if } I_{k,j} \in \{0, 1\}, \\
0, & \text{ if } I_{k,j}=\text{NA}.\\
\end{array}
\right. 
$$

The second task is a regression for the non-zero incremental payments $Y_{k,j}$. The regression loss is defined by a function $f$ that can either be the squared error $f(Y^{*},\hat{Y}^{*})=f_{SE}(Y^{*},\hat{Y}^{*})=(Y^{*}-\hat{Y}^{*})^2$ or the absolute error  $f(Y^{*},\hat{Y}^{*})=f_{AE}(Y^{*},\hat{Y}^{*})=|Y^{*}-\hat{Y}^{*}|$. The choice of function $f$ depends on the dataset and the dispersion of payment distribution. If the payments were only positive, we could consider, e.g., the gamma deviance, but here we allow for recoveries. For a given mini-batch, we evaluate the regression loss $RL$ as a weighted sum of development period loss functions $f$, with
\begin{align}
RL_f\left(\boldsymbol{\hat{Y}^{*}}\right) &= \sum_{j=2}^{n}  \frac{\sum_{i=1}^{b}\tilde{\delta}_{i,j}}{\sum_{h=2}^{n} \sum_{i=1}^{b} \tilde{\delta}_{i,h}} \left\{\sum_{k=1}^{b} \frac{\tilde{\delta}_{k,j}}{\sum_{i=1}^{b}\tilde{\delta}_{i,j}} \ f\left( Y_{k,j}^{*},\widehat{Y}_{k,j}^{*}\right)\right\} \notag\\
&= \frac{1}{\sum_{h=2}^{n} \sum_{i=1}^{b} \tilde{\delta}_{i,h}} \sum_{j=2}^{n}   \sum_{k=1}^{b} \tilde{\delta}_{k,j} \ f\left( Y_{k,j}^{*},\widehat{Y}_{k,j}^{*}\right), \label{RL}
\end{align}
where $\boldsymbol{\hat{Y}^{*}}=\{\hat{Y}_{k,j}^{*}:k=1, \ldots, b, j=2, \ldots, n\}$ and to only use non-zero observed targets, we set
$$\tilde{\delta}_{k,j} = \left\{
\begin{array}{rl}
0, & \text{if } Y_{k,j}^{*} \in \{-\mu_{\mathcal{T}}/\sigma_{\mathcal{T}}, \text{NA}\}, \\
1, & \text{otherwise}. \\
\end{array}
\right. 
$$
We recall that $Y_{k,j}^{*} =-\mu_{\mathcal{T}}/\sigma_{\mathcal{T}}$ correponds to the case of a zero payment, when  $I_{k,j} =0$. 

Our LSTM can learn well both tasks since they share informative features, which may induce more robust regularization and boost performance. However, for training, we must define a single loss function in terms of losses~(\ref{CE}) and (\ref{RL}). Multi-task network loss functions are often taken to be linear in the individual task losses. For example, the loss in \cite{gabrielli2020individual} is a weighted sum of the individual losses such that they lie on the same scale. Nevertheless, when learning  regression and classification simultaneously, we should ensure that both tasks are given equal consideration to prevent the easier one from dominating. To this end, several robust automatic loss balancing techniques were proposed; for an overview of multi-task learning approaches, see, e.g., \cite{ruder2017overview}.  

For training our network, we use the task-dependent uncertainty weighting technique proposed in \cite{kendall2018multi}. This approach allows the optimal weighting of losses with different units or scales and can be applied to combine classification and regression tasks. The multi-task optimization objective is defined as a Gaussian likelihood with task-dependent uncertainty. \cite{kendall2018multi} illustrated the performance of their loss balancing technique through a multi-task network learning visual scene, which outperformed a separate training of the same tasks. 

In several training experiments, our multi-task LSTM network outperforms separate learning of the classification and regression tasks. We observed a difference in optimization challenge and measurement scale between the losses~\eqref{CE} and~\eqref{RL}. In addition to using the task-dependent uncertainty, we introduce a scale hyper-parameter $\alpha$ applied to the classification loss to reduce the difference with the regression loss and improve tasks learning. Hence, the Gaussian likelihood function is 
\begin{equation}
\mathcal{L}(\boldsymbol{\hat{Y}^{*}},\ \boldsymbol{\hat{p}}) =\frac{1}{\sigma_1^2}\times RL_f(\boldsymbol{\hat{Y}^{*}}) + \frac{\alpha}{\sigma_2^2}\times CE\left(\boldsymbol{\hat{p}}\right)  + \log \sigma_1^2 +\log \sigma_2^2,   \label{eq:balanced_loss}
\end{equation}
with $\sigma_1^2$ and $\sigma_2^2$ corresponding to the regression and classification uncertainties, respectively. Note that we could set $\alpha=1$ if the tasks loss functions lie on similar scales. The optimization problem can be expressed, in terms of the LSTM network weights vector $\boldsymbol{W}$, as 
\begin{equation*}
\underset{\boldsymbol{W}, \sigma_1,\sigma_2}{\text{argmin}} \ \mathcal{L}(\boldsymbol{\hat{Y}^{*}},\ \boldsymbol{\hat{p}}).
\end{equation*}

We now explain several strategies used to improve the network training. Moving forward in periods as illustrated in Figure~\ref{LSTM_model}, we apply a procedure inspired by the teacher forcing technique to accelerate the training process (see, e.g., \cite{williams1989learning,drossos2019language}). According to a probability function that is exponentially increasing over epochs, at each time period we randomly choose whether to replace the observed pair $(I_{k,j},\ Y_{k,j}^{*})$ within the input $\boldsymbol{X}_{k,j}$ with the predictions from the previous period $(\hat{p}_{k,j},\ \hat{p}_{k,j}\times\hat{Y}_{k,j}^{*})$. This improves the ability of the network to predict correctly late periods even when there are imprecise inputs in early periods. Through several training experiments, we find that replacing the observed values with the expected ones outperforms the replacement with simulated values. Orange dotted lines in Figure~\ref{LSTM_model} represent the replacement procedure, which is similar to the missing values imputation. We use stochastic gradient descent with an initial learning rate $l_r$, that we drop by a factor when there is no improvement in the validation loss for $r_p$ epochs. The training process is stopped early when the validation loss does not improve for $t_{es}$ epochs, and then we obtain the optimal weights.

Our network loss function~(\ref{eq:balanced_loss}) does not have a simple interpretation. We are thus interested in some metrics to evaluate the model's performance. For classification, we use the ROC (Receiver Operating Characteristic) curve with observations grouped by development period and the associated AUROC (Area Under the ROC). The higher the AUROC, the better the model distinguishes between having a payment or not. Other metrics are presented in Sections~\ref{subsec:resultat_sim} and \ref{subsec:resultat_real}.

\section{Simulated data}\label{sec_simulated}
	
In this section, we evaluate our individual loss reserving model performance on fully developed micro-level simulated data.  This allows us to compare the predictive outstanding loss estimates with their actual values. One advantage of using simulated data is research reproducibility; the complete code for this case study is available from the authors upon request. Moreover, we conduct a comparative analysis between our model and the aggregate chain-ladder approach. 

\subsection{Data description}
	
Using NNs, \cite{gabrielli2018individual} developed a stochastic generator of non-life insurance individual claim histories, based on a real property and casualty insurance dataset. The exact setup used for the generator is provided in Appendix \ref{subsec:appA}.

The simulated dataset contains $998,807$ claims with occurrence dates between years 1994 and 2005. For claim $k$, several static characteristics are contained in vector $\boldsymbol{S}_k$, for instance, claim number, accident year, claimant age, line of business, and injured body part. Through the yearly claim development, we know the incremental payments and indicators denoting whether the claim is open or closed. These two dynamic informations are reported at the end of each development year $j=1,\ldots, 12$, and claims are assumed to be fully developed after $n=12$ years. 

For our reserving model, we focus on the predictions of the occurrence and amounts of incremental payments (or recoveries). Hence, we drop claim status and keep the payment information within the dynamic vectors, with $\boldsymbol{D}_{k,j}=\{ j,\ r_{k,j},\ I_{k,j},\ Y_{k,j}^{*}\}$, for $j=1, \ldots, t_k$, and $k=1,\ldots,m$. However, we recall that it is possible to add a learning task to our NN to predict claim status. Table~\ref{tab:variables_sim_dataset} lists the static and dynamic variables used in our reserving model and their pre-processing. Note that the simulated claims have some realistic features such as negative payments representing recoveries (0.3\%), late reporting, reopenings (0.3\%), and settlement at zero (28\%). However, the data contain only few covariates to properly describe the characteristics of claims, which may harm the performance of micro-level reserving techniques.
	
\begin{table}[t]
	\centering
	\caption{List of simulated dataset features and their pre-processing.}\label{tab:variables_sim_dataset}
	\begin{tabular}{clc} \hline
	\textbf{Type} & \hspace{0.8cm}Features & Pre-processing \\ \hline
		\multirow{7}{*}{Static} & Accident year  &  Expressed in number of years from 1994\\
		& Reporting delay &   --- \\
		& Claimant age & Scaled to $[0, 1]$ \\
		& Line of business &   4 categories indexed \\
		& Claim code &  51 categories indexed\\
		& Injured part &  46 categories indexed \\ \hline
		Dynamic & Incremental yearly  payment  &  
		Centered and scaled as in Eq. \eqref{transform} \\ \hline
	\end{tabular}
\end{table}

We set $T^*=2005$ as the evaluation year. To train our reserving model, we keep only claims reported before $T^*$ and hide payments made beyond $T^{*}$. Hence, we retain $m=991,904$ claims leading to a deletion of only 0.7\% of them.  This small percentage is due to the fact that 92\% of claims are reported during their first development year. Also, claims close on average after 1.5 year. Therefore, we expect to forecast few future non-zero outstanding payments. We stratify the claims into training (60\%), validation (20\%), and testing (20\%) datasets. The hidden payments associated to each of the three datasets made after $T^*$ are used later to assess the network's performance. 

\begin{figure}[t]
	\centering
	\begin{subfigure}{0.49\textwidth}
		\includegraphics[scale=0.56]{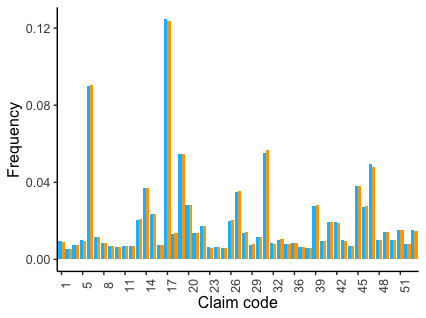}
		\subcaption{}\label{cc}
	\end{subfigure}
	\begin{subfigure}{0.49\textwidth}
		\includegraphics[scale=0.56]{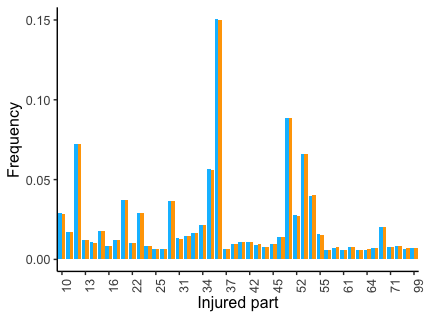}
		\caption{}\label{inj_part}
	\end{subfigure}
	\caption{Frequency of claim code (a) and injured part (b) for training (blue) and validation (orange) sets.}\label{fig:freq_categorical_sim}
\end{figure}

We carried out the same preliminary analysis on both training and validation datasets to assess the split homogeneity.  Among the available static information in $\{\boldsymbol{S}_{k}:k=1, \ldots, m\}$, we analyze the categorical covariates to index their categories and set the hyper-parameters of the embedding layers $\ell_1$, $\ell_2$ and $\ell_3$. Figure~\ref{fig:freq_categorical_sim}(\subref{cc}) depicts the frequency of claim code. We identify 51 categories with labels in $\{1, \ldots, 53\}$. For the injured body part, we observe 46 categories in the barplot illustrated in Figure~\ref{fig:freq_categorical_sim}(\subref{inj_part}), with labels within $\{10, \ldots, 99\}$.  Both figures show almost the same probability distribution for each variable for training and validation datasets, which confirm the similarity of the two datasets structures. The third categorical variable is the line of business with four categories and empirical frequency almost identical to the probability distribution given in Table~\ref{tab:para_generator}.

\subsection{Model training}\label{subsec:train_sim}

Due to the ultimate development period being $12$ years, we set $n=12$ in our LSTM network depicted in Figure~\ref{LSTM_model} and outputting the set of pairs  $\{(\hat{p}_{k,j}, \hat{Y}^{*}_{k,j}), j=2, \ldots, 12\}$. Since we do not observe large claims or high variance for incremental payments and we model recoveries, we choose the squared error function $f=f_{SE}$ in regression loss~(\ref{RL}) and LSTM network loss function (\ref{eq:balanced_loss}). After the pre-processing of variables  described in Table~\ref{tab:variables_sim_dataset}, we train our net using the training dataset. We perform several experiments to set the hyper-parameters on the validation dataset. The optimal model performance is obtained with the hyper-parameter combination given in Table~\ref{tab:hyppara_sim}.

\begin{table}[t]
	\centering
	\caption{Hyper-parameters 
		used to train the LSTM network on the simulated dataset.}\label{tab:hyppara_sim}
	\begin{tabular}{l c | l c } \hline
		Hyper-parameter & Value & Hyper-parameter & Value \\ \hline
		Context size $c$ & 32 & Hidden size & 128 \\
		Batch size $d$ & 2048 & 
		Learning rate $l_r$ & 0.05 \\
		Reduce on plateau $r_p$& 10 epochs &
		Early stopping $t_{es}$ & 15 epochs \\ \hline
	\end{tabular}
\end{table}

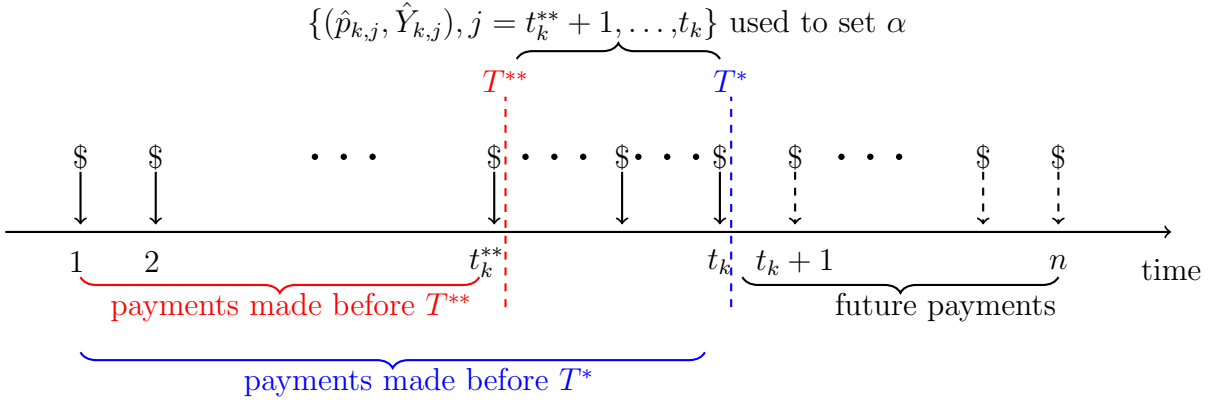
\begin{figure}[t]
	\centering
	\begin{tikzpicture}[scale=1]

	\draw [thick, ->] (1,0.8) -- (1,0.1);
	\node[align=center] at (1,1) {\$};
	\node[align=center] at (0.95,-.4) {1};
	\draw [thick, ->] (2,0.8) -- (2,0.1);
	
	\node[align=center] at (1.95,-.4) {2};
	\node[align=center] at (2,1) {\$};
	
	\node[align=center] at (4.5,1) {\huge{$\ldots$}};
	\node[align=center] at (6.5,1) {\$};
	\draw [thick, ->] (6.5,0.8) -- (6.5,0.1);
	
	\node[align=center] at (8.2,1) {\$};
	\draw [thick, ->] (8.2,0.8) -- (8.2,0.1);
	\node[align=center] at (7.3,1) {\huge{$\ldots$}};
	\node[align=center] at (9.5,1) {\$};
	\draw [thick, ->] (9.5,0.8) -- (9.5,0.1);
	\node[align=center] at (9.5,-.4) {$t_k$};
	\node[align=center] at (8.8,1) {\huge{$\ldots$}};
	
	\node[align=center] at (6.4,-.4) {$t_{k}^{**}$};
	
	\draw[thick,red,decorate,decoration={brace,amplitude=6pt,raise=0pt,mirror}] (1,-.6) -- (6.3,-.6);
	\node[align=center,red] at (3.8,-1) {payments made before $T^{**}$};

	\draw[thick,decorate,decoration={brace,amplitude=6pt,raise=0pt,mirror}, blue] (1,-1.6) -- (9.3,-1.6);
	\node[align=center, blue] at (5.5,-2) {payments made before $T^*$};
	
	\node[align=center,red] at (6.65,2) {$T^{**}$};
	\draw [thick, dashed,red] (6.65,-1) -- (6.65,1.8);
	
	\node[align=center, blue] at (9.65,2) {$T^{*}$};
	\draw [thick, dashed, blue] (9.65,-1) -- (9.65,1.8);

	\draw [thick, ->, dashed] (10.5,0.8) -- (10.5,0.1);
	\node[align=center] at (10.5,1) {\$};
	\node[align=center] at (10.5,-0.4) {$t_{k}+1$};
	\node[align=center] at (11.5,1) {\huge{$\ldots$}};
	\draw [thick, ->, dashed] (13,0.8) -- (13,0.1);
	\node[align=center] at (13,1) {\$};
	\node[align=center ] at (14,-0.4) {$n$};
	\node[align=center] at (14,1) {\$};
	\draw [thick, ->, dashed ] (14,0.8) -- (14,0.1);

	\draw[thick,decorate,decoration={brace,amplitude=6pt,raise=0pt,mirror}] (9.8,-0.6) -- (14,-0.6);
	\node[align=center] at (12.5,-1) {future payments};
	
	\draw [thick,->] (0,0) -- (15.5,0);
	\node[align=center] at (15.5,-0.5) {time};

	\draw[thick,decorate,decoration={brace,amplitude=6pt,raise=0pt}] (6.8,2.3) -- (9.5,2.3);
	\node[align=center] at (8,2.8) { $\{(\hat{p}_{k,j}, \hat{Y}_{k,j}), j=t_{k}^{**}+1,\ldots,t_k\}$ used to set $\alpha$};
	\end{tikzpicture}
	\caption{Development of claim $k$ over $n$ equal periods. }\label{fig:shifted_eval_date}
\end{figure}

To determine the scale hyper-parameter $\alpha$ in loss function (\ref{eq:balanced_loss}), we train our network with $\alpha \in \{0.1, 0.2, \dots, 2.5\}$. The choice of this grid is based on our training experiments. To set the best value, we need to compare the predictions with the actual payments while using only the information known at evaluation time $T^*=2005$. We explain in the following how to do so. We set the end of the year 2003 as our reference date, denoted $T^{**}$ and,  for claim $k$, we define $t_{k}^{**}=t_k-(T^{*}-T^{**})$ as the number of development periods observed at $T^{**}$, as illustrated in Figure~\ref{fig:shifted_eval_date}. In the new validation dataset $\mathcal{V}$, we select the claims for which at least one development year is observed at $T^{**}$, i.e., $t_k^{**}>0$, leading to a loss of 18\% of the observations. 

Our choice of hyper-parameter $\alpha$ is based on two performance metrics. First, the ratio of the aggregate predicted over the aggregate observed between $T^{**}$ and $T^{*}$ is
\begin{equation*}
RR(T^{**},T^*)=\frac{\sum_{k\in \mathcal{V}}\sum_{j=t_{k}^{**}+1}^{t_k} \mathbbm{1}_{\{t_k^{**}>0\} } \times \hat{p}_{k,j} \times \hat{Y}_{k,j} }{\sum_{k\in \mathcal{V}}\sum_{j=t_{k}^{**}+1}^{t_k}  \mathbbm{1}_{\{t_k^{**}>0\} } \times  I_{k,j} \times Y_{k,j} }.
\end{equation*}
The second is the ratio of the aggregate ultimate assuming the ultimate time is $T^*$, given by
\begin{equation*}
	RU(T^{**},T^*)=\frac{\sum_{k\in \mathcal{V}} \mathbbm{1}_{\{t_k^{**}>0\}} \left(\sum_{j=1}^{t_{k}^{**}} I_{k,j} \times Y_{k,j}+ \sum_{j=t_{k}^{**}+1}^{t_k} \hat{p}_{k,j} \times \hat{Y}_{k,j}\right)}{\sum_{k\in \mathcal{V}} \sum_{j=1}^{t_k} \mathbbm{1}_{\{t_k^{**}>0\}} \times I_{k,j} \times Y_{k,j}}.
\end{equation*}
In the training experiments, we observe that both ratios have a non-monotonic relationship with~$\alpha$. LSTMs trained with $\alpha \in\{0.2, 0.6, 0.8\}$ outperform the other networks on the validation subset. The ratios given in Table~\ref{tab:ratio_ult_sim} are closest to one in the case $\alpha=0.2$, which is what we select.
\begin{table}[t]
	\centering
	\caption{Ratios on the simulated validation dataset for the selection of $\alpha$.}\label{tab:ratio_ult_sim}
	\begin{tabular}{lccc} \hline
		\hspace{1cm}$\alpha$ & 0.2 & 0.6 & 0.8  \\ \hline
	 $RR(2003,2005)$ & 0.9556  &  0.9145& 0.9413 \\
	 $RU(2003,2005)$ &  0.9950 & 0.9904 &  0.9934 \\ \hline
	\end{tabular} 
\end{table}

\subsection{LSTM results and comparison with chain-ladder}\label{subsec:resultat_sim}

To assess our LSTM network efficiency, we use the validation and testing datasets. The latter has not contributed to the training process in any way, and the network will see it for the first time. In the following, we analyze the prediction accuracy at both the aggregate and individual levels. Also, we compare the results of the LSTM with those of the aggregate chain-ladder model. 

Using the payments that were previously hidden and $\mathcal{D}$ being either the validation or testing dataset, we evaluate the ratios of the aggregate reserve and ultimate given, respectively, by 
\begin{equation*}
\frac{\sum_{k\in \mathcal{D}}\sum_{j=t_{k}+1}^{n} \hat{p}_{k,j} \times \hat{Y}_{k,j} }{\sum_{k\in \mathcal{D}}\sum_{j=t_{k}+1}^{n} I_{k,j} \times Y_{k,j} } \qquad \text{   and  } \qquad \frac{\sum_{k\in \mathcal{D}} \left(\sum_{j=1}^{t_{k}} I_{k,j} \times Y_{k,j}+ \sum_{j=t_{k}+1}^{n} \hat{p}_{k,j} \times \hat{Y}_{k,j}\right)}{\sum_{k\in \mathcal{D}} \sum_{j=1}^{n}I_{k,j} \times Y_{k,j}}.
\end{equation*}
Note that actual reserve refers to the actual outstanding payments between $T^{*}$ and $n$, while actual ultimate corresponds to the observed total paid over $n$ periods. Table~\ref{tab:ratio_test_sim} presents the ratios evaluated with the predictions of the LSTM and chain-ladder. For the validation dataset, we obtain very similar results with both models. For the testing dataset, the chain-ladder underestimates the reserve by 2\% while our model overstates it by 3.9\%. Both models perform well on the ultimate with ratios very close to one. However, the LSTM has the advantage of producing individual predictions. Moreover, the characteristics of the simulated data are very close to the chain-ladder hypotheses, which explains these results. Section~\ref{sec_real_data} presents a case study with real data where we observe a more important difference between the performance of these models.

\begin{table}[t]
	\centering
	\caption{Aggregate ratios using $\alpha=0.2$ on the simulated datasets.  }\label{tab:ratio_test_sim}
	\begin{tabular}{l|cc|cc}\hline
		\multirow{2}{*}{Datasets} & \multicolumn{2}{c}{Ratios of aggregate reserve } &  \multicolumn{2}{|c}{Ratios of aggregate  ultimate}\\  
		&  LSTM  & Chain-ladder &  LSTM & Chain-ladder   \\ \hline
		Validation & 1.0271 & 1.0245 & 1.0033 & 1.0030  \\ 
		Testing & 1.0385 &  0.9806  & 1.0048 & 0.9975 \\ \hline
	\end{tabular} 
\end{table}

\begin{figure}[t]
	\centering
	\includegraphics[width=16cm, height=4cm]{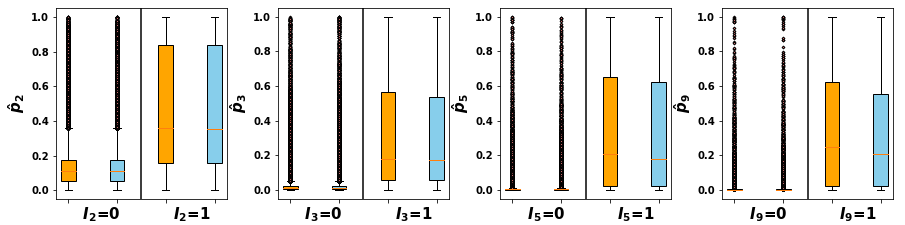}
	\caption{Boxplots of the predicted probability $\hat{p}_{k,j}$ in terms of the observed $I_{k,j}$, for $j\in\{2, 3, 5, 9\}$ and for validation (orange) and testing (blue) simulated datasets.}\label{fig:boxplot_sim}
\end{figure} 

\begin{figure}[b]
	\centering
	\includegraphics[scale=0.535]{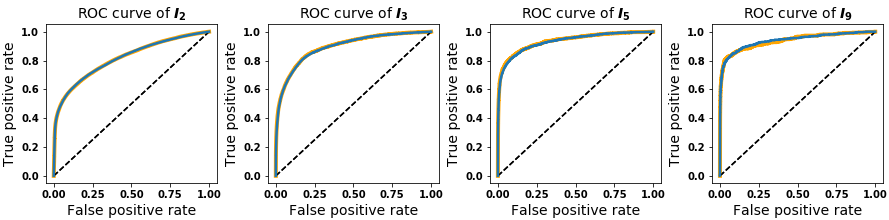}
	\caption{ ROC curves on the payment indicator classification, for periods $j\in \{2, 3, 5, 9\}$ for validation (orange) and testing (blue) simulated datasets.}\label{fig:ROC_sim}
\end{figure}

\begin{figure}[h]
	\centering
	\includegraphics[scale=0.45]{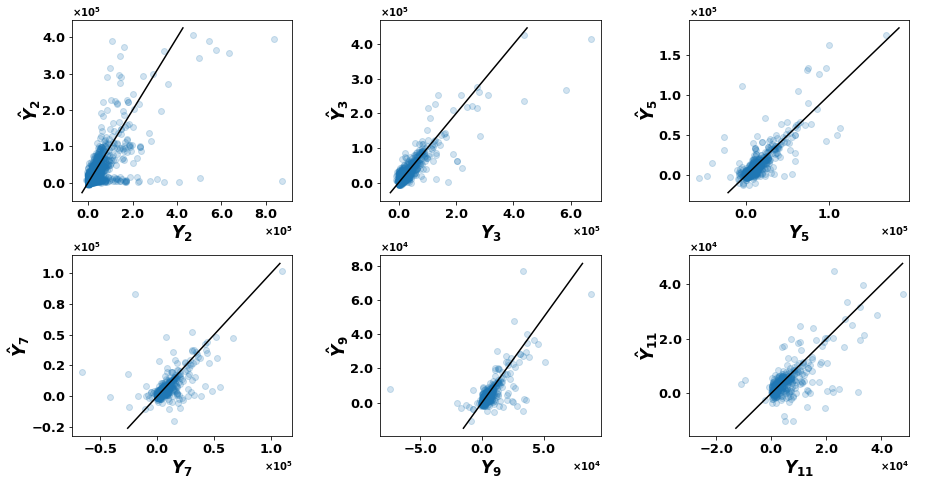}
	\caption{ Predicted payments in function of observed non-zero payments, for periods $j\in\{2, 3, 5, 7, 9, 11\}$ for the simulated testing dataset.}\label{fig:predicted_payments_sim}
\end{figure}

\begin{figure}[t]
	\centering
	\includegraphics[scale=0.405]{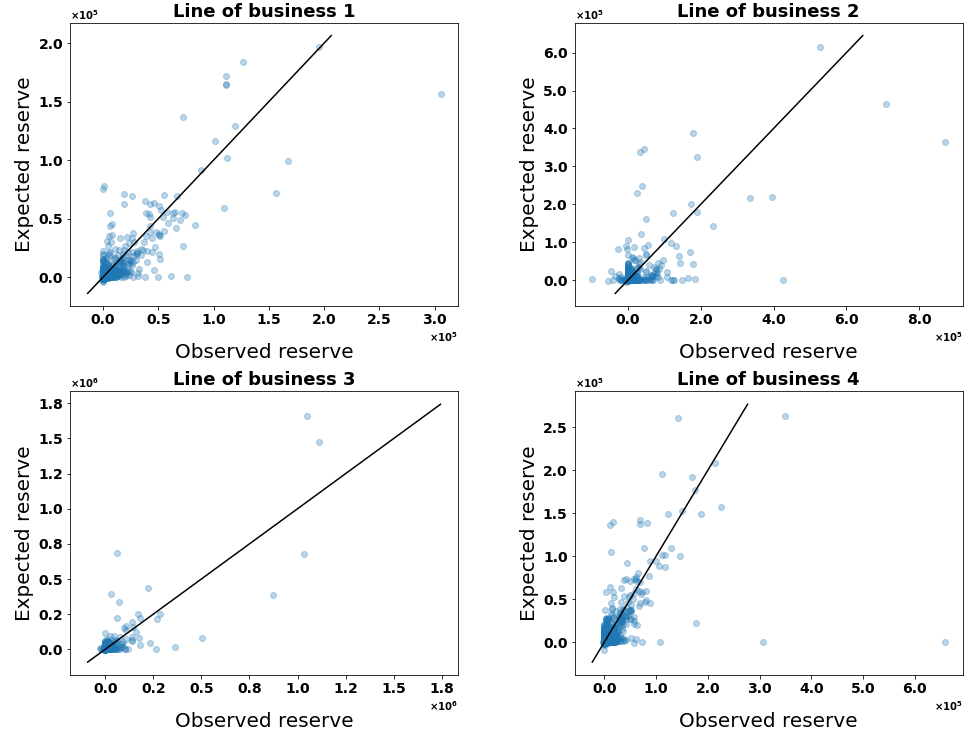}
	\caption{ Expected versus observed individual reserves for simulated testing dataset per line of business.}\label{fig:reserve_per_lob}
\end{figure}

At the individual level, we proceed to investigate the performance on the classification and regression tasks. Figure~\ref{fig:boxplot_sim} depicts the boxplot of the predicted probability of payment in terms of the observed indicator, for periods $j\in \{2, 3, 5, 9\}$ from left to right, for validation and testing datasets. As the claim development progresses, we better predict $\Pr(I_{k,j}=0)$ compared to $\Pr(I_{k,j}=1)$. This is due to the short average lifetime of claims, 1.5 year, which implies a large proportion of zero values for the indicator $\{I_{k,j}, j=3, \ldots, n\}$. Figure~\ref{fig:ROC_sim} illustrates the ROC curves for periods $j\in \{2, 3, 5, 9\}$. The closer the curve is to point $(0, 1)$, the better the model distinguishes between categories. We once again observe an improvement of LSTM probability predictions over periods.

We illustrate in Figure~\ref{fig:predicted_payments_sim} the scatterplots of incremental predicted payments, only for non-zero observed payments as that is how we train the regression task. We observe that the majority of points are close to the diagonal, which reflects the precision of predictions. In period $j=2$, some predictions are close to zero whatever the value of the observed; this is expected as, in that case, the LSTM knows only one development period. The predictions improve as development progresses. From the LSTM output $\{(\hat{p}_{k,j}, \hat{Y}_{k,j}^{*}),j=2,\ldots,n\}$, we compute the expected individual reserve $\hat{R}_{k}=\sum_{j=t_k+1}^{n} \hat{p}_{k,j}\times\hat{Y}_{k,j}$.  Figure~\ref{fig:reserve_per_lob} presents the scatterplots of $\hat{R}_{k}$ in terms of the observed reserves $R_k$ per line of business. In general, the predictions of our network are satisfying. We note a few cases of over or under-estimation and a difference in claim sizes by line of business.

%\clearpage

\section{Real dataset}\label{sec_real_data}

In this section, we illustrate our procedure on a real individual claims dataset provided by a large Canadian insurance company. First, we provide a brief description of the data. Then, we explain how we adapt our reserving model to the data and we describe the training process. As some extreme payments are hard to capture with NNs, we propose a large claim forecasting procedure. Our reserve estimates are compared with those obtained with the aggregate chain-ladder method.

\subsection{Data description}\label{subsec:data_analyse}

We consider a real general insurance individual claims dataset from a Canadian insurance company. The detailed dataset relates to auto insurance policies and contains bodily injury claims that occurred from January $1^{\text{st}}$, 2006 to August $31^{\text{st}}$, 2010. There are $m=53,677$ claimants associated to $41,916$ distinct claims. Note that we implicitly consider the dependence between claimants of the same claim through common covariates such as vehicle characteristics, and number of claimants.

From the reporting date, each claim's development is described through a monthly evolution of its individual information, divided into two categories. First, the static characteristics of claimant file $k$ are contained in a vector $\boldsymbol{S}_k$, for $k=1,\ldots,m$. This may include information on the claim (date, location, driver experience), the claimant (age, gender), and the vehicle (age, number of cylinders). The second category groups dynamic information, for instance, the number of claimants, the nature and number of injuries and the payment. Tables~\ref{tab:list_static_real} and \ref{tab:list_dynamic_real} list the static and dynamic features used in our model.

First, we pre-process the real dataset. We correct wrong entries, e.g., an accident date greater than the reporting date. To deal with missing values, we apply a multiple imputation by chained equations (MICE), using the $\textsf{R}$ package \textit{mice} (\cite{van2015package}). For categorical variables with many modalities (e.g., postal code), this procedure is computationally intensive. We hence apply the MICE approach to only ten of the 17 available variables with missing values, and we drop the other ones. See Table~\ref{tab:list_static_real} for a description of the variable pre-processing. 

We set the evaluation date $T^*$ as August $31^{\text{st}}$, 2010.  By analyzing the development of claimant files closed as of $T^*$, we observe that their lifetime is on average three months and they have a 0.3\% probability of reopening. The probability of a claimant file lasting longer than 45 months is very low. Therefore, we are interested in studying only the first 45 months of claimant file development, i.e., we set $n = 45$ for our LSTM. This development censorship allows to have enough observations for the network to train on the last period predictions. Also, we hide all information observed beyond $T^*$, then stratify the dataset into training (60\%), validation (20\%), and testing (20\%) datasets. The same exploratory analyses are carried out on the training and validation datasets, and the dataset split seems homogeneous, as seen in Figure~\ref{graphe_analyse}.
 
\begin{figure}[t]
	\centering
	\begin{subfigure}{0.48\textwidth}
		\includegraphics[scale=0.5]{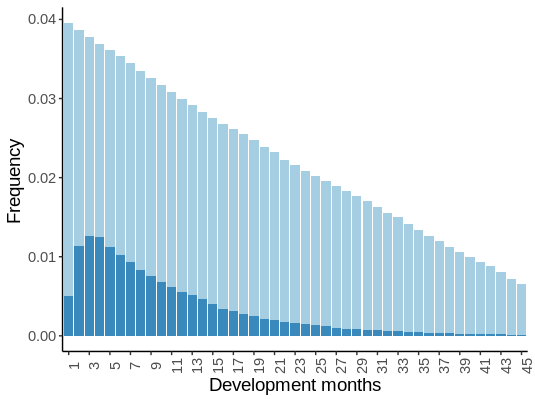}
		\caption{}
	\end{subfigure}
	\begin{subfigure}{0.48\textwidth}
		\includegraphics[scale=0.5]{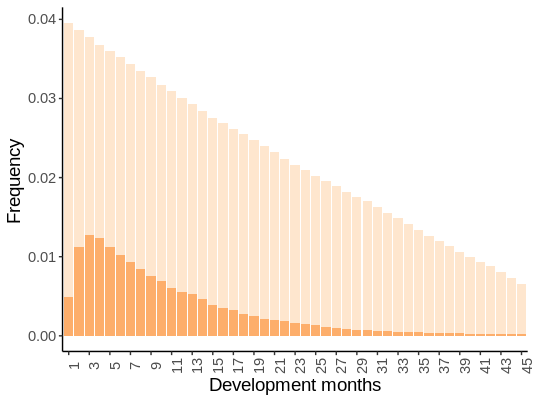}
		\caption{}
	\end{subfigure}
	\caption{Frequency of observations without payment (light color) or non-zero payment (dark color) for training (a) and validation (b) datasets by development month.}\label{freq_observertion}
\end{figure}

The real dataset has several features in common with the simulated dataset discussed in Section~\ref{sec_simulated}. In both, we observe reopenings, and around 30\% of claimants files settled without any payment even though they may remain open for several periods. However, compared to the simulated dataset, the real data has the advantage of having a monthly (rather than yearly) detailed development with a greater number of dynamic features (other than incremental payments). If we wanted to include the future development of these features in the network, it would imply an additionnal task. To avoid adding complexity, we consider the evolution of dynamic claimant file characteristics other than the payments until $T^*$ and then keep the last observed value for the remaining development months. Thus, we take advantage of this information to improve the network predictions for the first development months.

The number of observed targets used to learn the classification and regression tasks differs and may impact the complexity of the optimization of functions~(\ref{CE}) and (\ref{RL}). For the training and validation datasets, Figure~\ref{freq_observertion} illustrates two frequency distributions. The first one, in light color, is related to observed zero incremental payments per development month. The second distribution, in dark, relates only to non-zero incremental payments among all observed payments employed to compute the regression loss~(\ref{RL}). The concatenation of the two colors corresponds to the distribution of observed indicators $ I_ {k, j} $, for $ j = 1, \ldots, t_k $, used to compute the classification loss~(\ref{CE}). There are fewer observations within both datasets to train the regression task than the classification. This is expected given the large percentage of claimant files settled without any payment.

 \begin{figure}[t]
	\centering
	\includegraphics[scale=0.55]{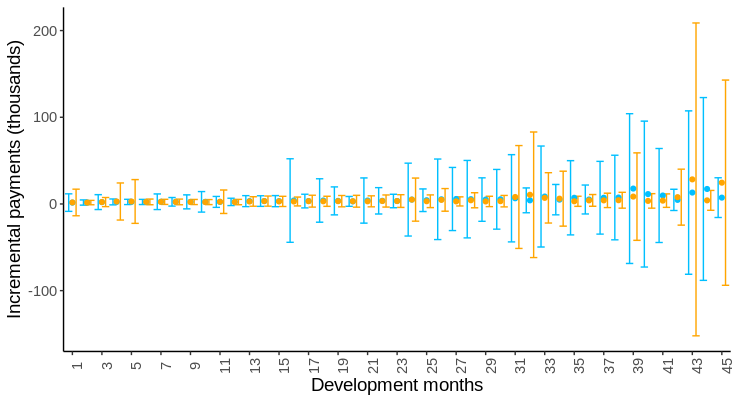}
	\caption{Mean and standard deviation of non-zero incremental payments at the evaluation date $T^*$ for training (blue) and validation (orange) datasets by development month. }\label{boxplot_payments}
\end{figure}

To investigate in details the incremental payments, we depict in Figure~\ref{boxplot_payments} error bars representing the standard deviation around the mean for the non-zero payments, per development month. We observe huge variances for some periods indicating the presence of extremely large payments. Recall that we did not observe extreme claims in the simulated dataset from Section~\ref{sec_simulated}.  We propose in Section~\ref{large_claims} an approach to handle large payments.

\subsection{Model for extreme payments}\label{large_claims}

Large or catastrophic claims are extreme events leading to very high payments and can be modeled using Extreme Value Theory (EVT). Given a high threshold $u>0$, EVT allows to model the conditional distribution of the excess loss over $u$ with a generalized Pareto distribution (GPD) as limiting distribution. For a general introduction to EVT, see, e.g., \cite{coles2001introduction}.

Using only the training dataset, we consider $\boldsymbol{Y}=\{Y_{k,j}:Y_{k,j}\neq 0 , j=1, \ldots, t_k, \ k=1, \ldots, m\}$ the vector of non-zero incremental payments known at the evaluation date. We aim to model the tail behavior of $\boldsymbol{Y}$ using EVT. We set a sufficiently high threshold $u$, and estimate the parameters of the GPD associated to the random excess  $(Y_{k,j}-u|Y_{k,j}>u)$.  Then, we verify that the empirical behavior of $\boldsymbol{Y}$ can be described by the selected GPD.

To choose a threshold $u$, we use the mean excess plot, and $u$ is selected as the lowest threshold for which the curve becomes approximately linear. Based on the detailed analysis provided in Appendix~\ref{subsec:appC}, we set $u = 32,000$. With the selected threshold, we estimate the GPD parameters  by maximum likelihood through a numerical optimization with the $\textsf{R}$ package \textit{ismev} (\cite{ismev}). We obtain shape $\hat{\sigma}=0.957851$ and scale $\hat{\lambda}=14,044$. Therefore, $\hat{E}[Y_{k,j}-u|Y_{k,j}>u]=\hat{\lambda}/(1-\hat{\sigma})=333,199$, i.e., given that a payment exceeds the threshold $u=32,000$, it will cost on average \$365,199. The goodness-of-fit of the estimated GPD is verified in Appendix \ref{subsec:appC}.

\begin{figure}[t]
	\centering
	\includegraphics[scale=0.5]{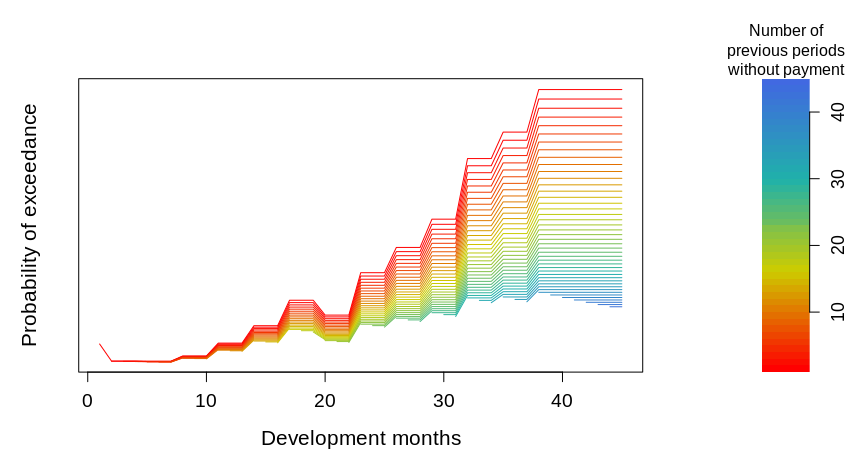}
	\caption{Relative predicted probabilities of large payment in terms of development month and number of previous periods without payment.}\label{fig:prob}
\end{figure}

Now that we have the expected excess amount, we move to accurately estimating the probability that a payment exceeds $u$. Note that we have only 0.5\% of available non-zero payments $\boldsymbol{Y}$ above the threshold $u=32,000$. In addition to the development period as a predictive variable, an analysis of the payment history, inspired by \cite{cote2021bayesian}, revealed that it is less likely to observe large payments after multiple consecutive periods without payment. We use a generalized linear model (GLM) for the probability of large payment knowing the development period and the number of previous months without payment. Given the few observations (sometimes less than five) exceeding $u$ for some development periods, we group months in quarter periods and also merge development months 39 to 45. We keep the first month apart given its particular behavior. Figure~\ref{fig:prob} depicts the predicted relative probabilities of large payment obtained with our GLM, per development month. The number of previous periods without payment is represented using the color scale; as it increases (colder color), the probability of extreme payment decreases. Fitted probabilities of extreme payments also increase with development month.

The large claims analysis developed in this section confirms our interpretation of Figure~\ref{boxplot_payments}. In addition, in the first training experiments on the real data, we found important discrepancies between the predicted and observed payments, most likely resulting from the considerable variance. Therefore, we propose to censor payments in order to reduce variance and improve learning. In that case, the LSTM prediction are interpreted as the conditional expected payment given that it does not exceed the threshold $u$, and expressed, for claimant file $k$ and period $j=t_k+1,\ldots,n$, by
\begin{align}
\hat{E}[Y_{k,j}\vert Y_{k,j}\leq u] &= \hat{\Pr}\left(I_{k,j}=1\right)\times \hat{E}[Y_{k,j}\vert Y_{k,j}\leq u, I_{k,j}=1] \notag \\
&= \hat{p}_{k,j} \times \hat{Y}_{k,j}. \label{formula:pred}
\end{align}

We now aim to investigate an adjustment to the network's prediction to take into account the possibility that the payment is large. For claimant file $k$ and period $j$, we have
\begin{align}
\hat{E}[Y_{k,j}] &= \hat{\Pr}\left(I_{k,j}=1\right)\times \hat{E}[Y_{k,j}\vert I_{k,j}=1] \notag \\&= \hat{p}_{k,j} \times \left\{\hat{\Pr}\left(Y_{k,j}>u\right) \hat{E}[Y_{k,j}\vert Y_{k,j}>u, I_{k,j}=1]+\hat{\Pr}\left(Y_{k,j}\leq u\right) \hat{E}[Y_{k,j}\vert Y_{k,j} \leq u, I_{k,j}=1]\right\}  \notag\\
&= \hat{p}_{k,j} \times \left[ \hat{\Pr}\left(Y_{k,j}>u\right) \left(u+\frac{\hat{\lambda}}{1-\hat{\sigma}}\right)+\{1-\hat{\Pr}\left(Y_{k,j}>u\right)\} \hat{Y}_{k,j}\right], \label{formula:large_claims}
\end{align}
with $\hat{\Pr}\left(Y_{k,j}>u\right)$ the probability of exceedance $u$ obtained with our GLM in function of the development period and the number of previous months without payment. The latter covariate is approximated based on the LSTM predictions as explained in Section~\ref{subsec:large_claim_res}. 

If we consider (\ref{formula:large_claims}) instead of (\ref{formula:pred}) for all forecasted payments, we may overstate the reserve as we illustrate in Section~\ref{subsec:resultat_real}. Through an in-depth analysis of the results obtained for the training and validation datasets, we propose applying the adjustment given by (\ref{formula:large_claims}) for selected observations with greater probability of being large if the network was fed with uncensored payments.  To do so, for the training dataset, we analyze the distribution of predictions $\hat{Y}_{k,j}$ of large censored payments $Y_{k,j}$, to choose a threshold~$\zeta$ and we formulate the incremental payment prediction as follows
\begin{equation}
\hat{E}[Y_{k,j} ] = \left\{
\begin{array}{lcr}
\hat{p}_{k,j} \times \hat{Y}_{k,j}, &  & \text{if } \hat{Y}_{i,j} \leq \zeta \\
\hat{p}_{k,j} \times \left[ \hat{\Pr}\left(Y_{k,j}>u\right) \left(u+\frac{\hat{\lambda}}{1-\hat{\sigma}}\right)+\{1-\hat{\Pr}\left(Y_{k,j}>u\right)\} \hat{Y}_{k,j}\right],  &  &  \text{otherwise}.  \\
\end{array}
\right. \label{formula:LSTM_pred}
\end{equation}

\cite{delong2020neural} also use a GPD for large claims with periodic parameter estimation. However, since their neural net is not recurrent, at each period, they randomly select a proportion of observations with a high propensity to generate large claims according to some key features and estimate the prediction outside the network. The choice of the percentage and key features is very sensitive and requires a good knowledge about the claims nature. In our recurrent network, the prediction process is linked from period to period and considers the claimant file past development. According to our observations, it is more efficient to set a threshold $\zeta$ instead of a percentage to adjust large claim predictions. Unlike \cite{delong2020neural}, we use a GLM to evaluate the probability of large claims based on some dynamic covariates.

\subsection{Model training}\label{subsec:train_real}

Using censored payments $\min\left(Y_{k,j}, \ u\right)$, we train our LSTM network to learn optimal weights and predict the individual reserve accurately. Due to the difference in the number of observations used to compute functions~(\ref{CE}) and (\ref{RL}), we note a greater difficulty to learn the regression task than the classification one during our training experiments. Indeed, throughout the first epochs, we may observe a large difference between a prediction $\hat{Y}_{k,j}$ and its target $Y_{k,j}$. To reduce the regression loss scale values and make them closer to cross-entropy values, we choose the absolute error function $f_{AE}$ instead of the squared error $f_{SE}$ in losses~(\ref{RL}) and (\ref{eq:balanced_loss}).

After several LSTM training experiments, we set the hyper-parameter combination, given in Table~\ref{tab_hyperpara}, that optimizes the model performance. Compared to the LSTM network fitted to the simulated dataset in Section~\ref{sec_simulated}, the real data has inputs $\boldsymbol{X}_{k,j}$, $j=1, \ldots, n-1$, of higher dimension due to a greater number of dynamic features. Thus, we increased the size of the hidden state $\hidden_{k,j} $, for $j=1, \ldots, n-1$, to avoid losing information and enhance network performance. Note that a higher dimension requires more memory during the training process. Moreover, according to the dataset characteristics presented in Section~\ref{subsec:data_analyse}, when the mini-batch size is too small, the network will see a non-zero payment with low probability, particularly for the last development months. Hence, we choose a batch size of 1024 to have enough targets to learn the regression task and enough mini-batches per epoch to train our model.

\begin{table}
	\centering
	\caption{Hyper-parameters 
		used to train the LSTM network on the real dataset.}\label{tab_hyperpara}
	\begin{tabular}{l c | l c }  \hline
		Hyper-parameter & Value & Hyper-parameter & Value \\ \hline
		Context size $c$ & 16 & Hidden size & 256 \\
		Batch size $d$ & 1024 & 
		Learning rate $l_r$ & 0.05 \\
		Reduce on plateau $r_p$& 10 epochs &
		Early stopping $t_{es}$ & 15 epochs \\ \hline
	\end{tabular}
\end{table}

To set the hyper-parameter $\alpha$  without using the information after the evaluation date $T^*$, i.e., August $31^{\text{st}}$, 2010, we follow the same approach used in Section~\ref{subsec:train_sim} on the simulated dataset. We set May $31^{\text{st}}$, 2010, as a reference date, denoted $T^{**}$. We select the claimant files from the validation dataset with at least three months of development observed at $T^{**}$ such that $t_k^{**}>0$, leading to a loss of 20\% of the observations. We evaluate the metric ratios  $RR(T^{**},T^*)$ and $RU(T^{**},T^*)$ on this validation subset. Again, our results show no monotonic relationship between these ratios and $\alpha$. LSTMs trained with either $\alpha$ equal to 0.5 or 1.5 lead to ratios very close to one and outperform the other trained networks. Both choices produce a very similar performance at the aggregate level. At the individual level, on the validation subset, we note an overestimation of the small actual outstanding payments for the model with $\alpha=1.5$. Thus, we choose $\alpha=0.5$.

\subsection{LSTM results with censored payments}\label{subsec:resultat_real}

\begin{table}[t]
	\centering
	\caption{Aggregate ratios based on censored payments, with $\alpha=0.5$.  }\label{tab:ratio_test_real}
	\begin{tabular}{l|cc|cc}\hline
		\multirow{2}{*}{Datasets} & \multicolumn{2}{c}{Ratios of aggregate reserve} &  \multicolumn{2}{|c}{Ratios of aggregate ultimate}\\  
		& LSTM   & Chain-ladder & LSTM   & Chain-ladder \\ \hline
		Validation & \textbf{1.0044} & 1.1124 & \textbf{1.0017} & 1.0379  \\ 
		Testing & \textbf{0.9970} & 1.1238 & \textbf{0.9988}& 1.0436 \\ \hline
	\end{tabular} 
\end{table}

\begin{figure}[t]
	\centering
	\includegraphics[scale=0.5]{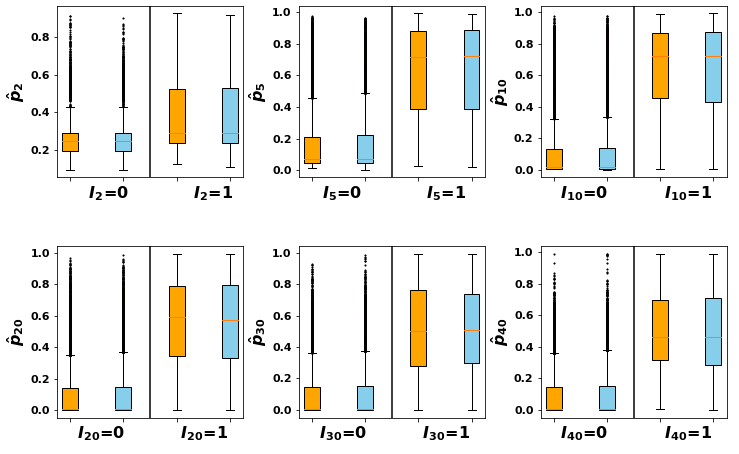}
	\caption{Boxplots of the predicted probability $\hat{p}_{k,j}$ in terms of the observed indicator $I_{k,j}$, for~$j\in\{2,5,10,20,30,40\}$ and for validation (orange) and testing (blue) real datasets.}\label{fig:boxplot_classification}
\end{figure}

In this section, we assess the efficiency of our LSTM network trained with censored payments based on the validation and testing datasets. First, we evaluate the accuracy of predictions at the aggregate level and then we compare it to the aggregate chain-ladder model. Second, at the individual level, we investigate the performance of the regression and classification tasks. Note that all computations in this section are made  only with payments censored at $u=32,000$.

Tables~\ref{tab:ratio_test_real} presents the ratios of the aggregate reserve and ultimate  obtained with both LSTM and chain-ladder. The expressions based on censored payments are given, respectively, by
\begin{equation*}
\frac{\sum_{k\in \mathcal{D}}\sum_{j=t_k+1}^{n} \hat{p}_{k,j} \times \hat{Y}_{k,j} }{\sum_{k\in \mathcal{D}}\sum_{j=t_k+1}^{n} I_{k,j} \times \min\left(Y_{k,j},u\right) } \quad \text{and} \quad \frac{\sum_{k\in \mathcal{D}} \left(\sum_{j=1}^{t_k} I_{k,j} \times \min\left(Y_{k,j},u\right)+ \sum_{j=t_k+1}^{n} \hat{p}_{k,j} \times \hat{Y}_{k,j}\right)}{\sum_{k\in \mathcal{D}} \sum_{j=1}^{n}I_{k,j} \times \min\left(Y_{k,j},u\right)}.
\end{equation*}
Note that the chain-ladder development factors are estimated using the censored training dataset. For validation and testing datasets, our network outperforms the chain-ladder with ratios very close to one. The chain-ladder overstates the reserve by more than 10\%. Thus, our model can better capture claim trends, which leads to a more accurate forecast using individual information. 

\begin{figure}[t]
	\centering
	\includegraphics[scale=0.52]{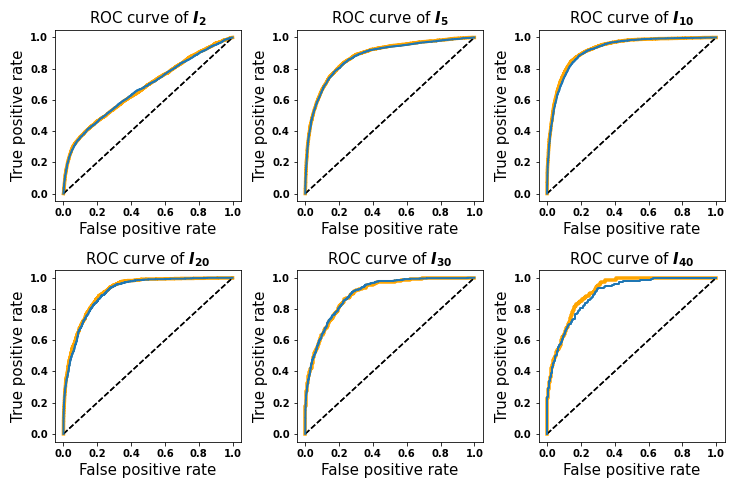}
	\caption{ ROC curves on the payment indicator classification, for periods $j\in \{2,5,10,20,30,40\}$ for validation (orange) and testing (blue) real datasets.}\label{fig:ROC}
\end{figure}

\begin{figure}[t]
	\centering
	\includegraphics[scale=0.45]{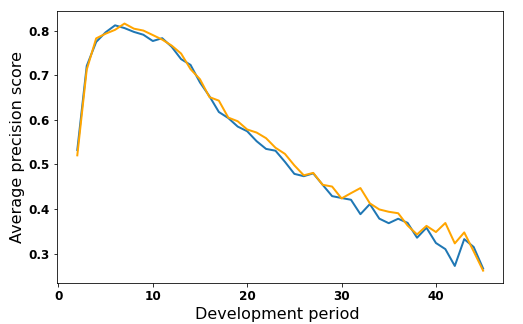}
	\caption{AUROC per development month for validation (orange) and testing (blue) real datasets.}\label{fig:AUROC}
\end{figure}

Figure~\ref{fig:boxplot_classification} presents boxplots of the predicted probability of payment, equivalently the probability that the indicator $I_{k,j}$ takes value 1 or 0 for periods $j\in \{2, 5,10,20,30,40\}$. For validation and testing datasets, we observe that we better predict $\Pr(I_{k,2}=0)$ compared to $\Pr(I_{k,2}=1)$.  As the claimant file process progresses and the network learns the development, we note an improvement in the prediction of probabilities. However, there are very few payments to be made in last development months leading to an unbalanced dataset with a large presence of zero values for indicators. This can bias the interpretation of the boxplots for last periods. 

The ROC curves for the classification are shown in Figure~\ref{fig:ROC} for periods $j\in\{2,5,10,20,30,40\}$, for both validation and testing datasets. As for Figure~\ref{fig:boxplot_classification}, we observe that the more the network knows the claimant file's history, the better is the prediction performance. Besides, to better evaluate the classification task without the impact of unbalanced observations, we use the AUROC per development period illustrated as a curve in Figure~\ref{fig:AUROC}.  We note an improvement during the first ten development periods, followed by a decrease in prediction accuracy. It is most probably due to the number of observations available for training. We can improve the network's performance by providing more observations, as in the simulated dataset studied in Section~\ref{sec_simulated}.

\begin{figure}[h]
	\centering
	\includegraphics[scale=0.45]{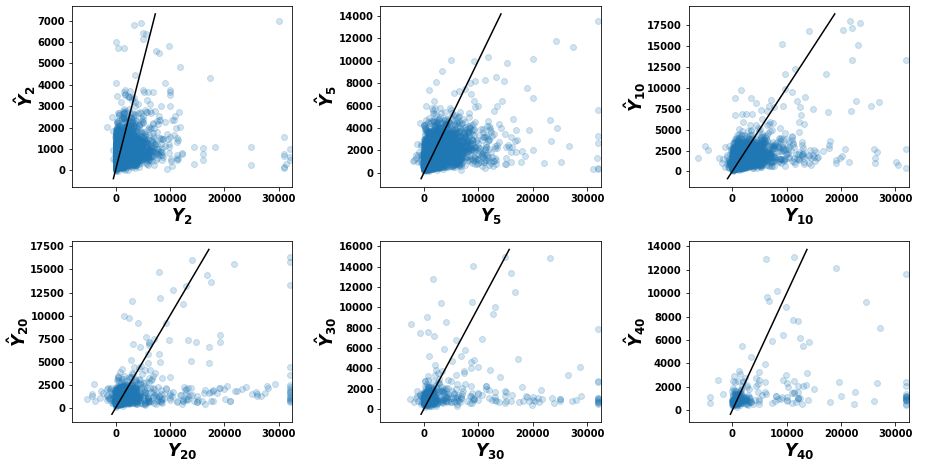}
	\caption{Predicted payments in function of observed non-zero payments for development month $j\in\{2,5,10,20,30,40\}$ for real testing dataset.}\label{fig:predicted_payments}
\end{figure}

Now, we move on to analyze the predicted payments. We illustrate in Figure~\ref{fig:predicted_payments} the scatterplots of incremental predicted payments for non-zero observed payments in the testing dataset. Moving forward in the development, we note an improvement in the prediction accuracy with points closer to the diagonal. On the other hand, beyond month 20, there is an underestimation of large observed payments, but, as discussed in Section~\ref{subsec:data_analyse}, few observations are available to train the regression task for the latest developments.

\begin{figure}[h]
	\centering
	\includegraphics[scale=0.44]{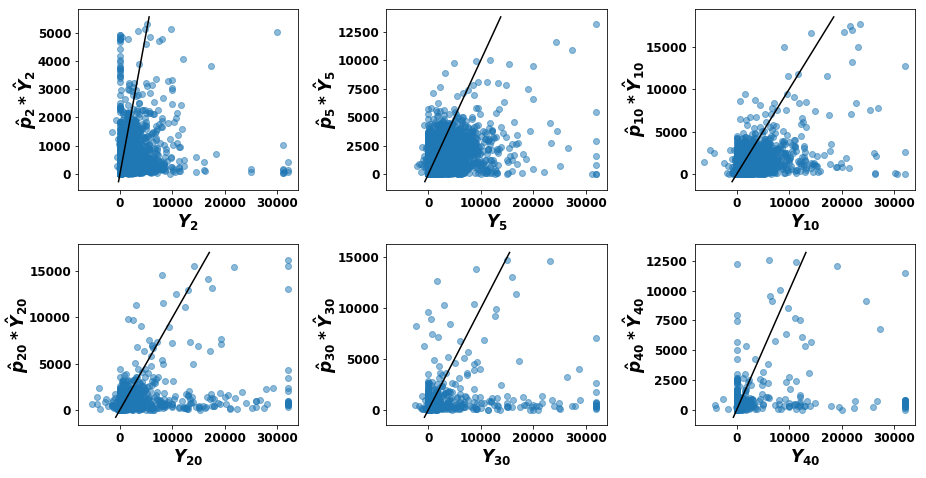}
	\caption{ Expected versus observed payments for month $j\in\{2,5,10,20,30,40\}$ in the real testing dataset.}\label{fig:Expected_values}
\end{figure}

\begin{figure}[h]
	\centering
	\includegraphics[scale=0.45]{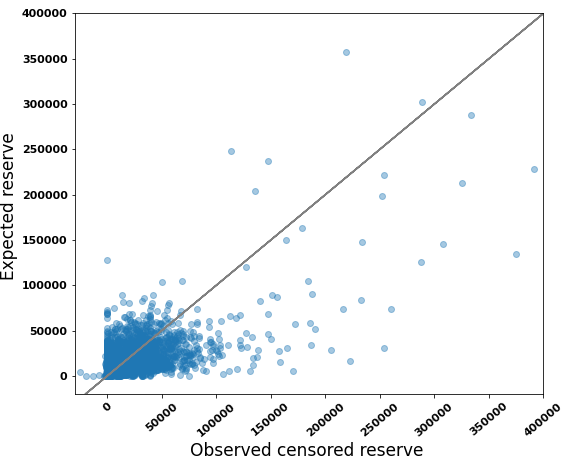}
	\caption{ Expected versus observed censored reserves for real testing dataset.}\label{fig:Expected_reserve}
\end{figure}

By multiplying the predictions of the regression and classification tasks, we obtain the expected payments $\hat{p}_{k,j}\times \hat{Y}_{k,j}$. They are presented for some periods in terms of the observed in Figure \ref{fig:Expected_values}. We note an overestimation of zero payments for period $j=2$. After month 20, we have an underestimation of the large observed payments and an overestimation of zero payments. By summing the expected incremental payments $\hat{p}_{k,j}\times\hat{Y}_{k,j}$ over future periods, we deduce the censored reserve $\hat{R}_{k}^{c}=\sum_{j=t_k+1}^{n} \hat{p}_{k,j}\times\hat{Y}_{k,j}$ for claimant file $k$. Figure~\ref{fig:Expected_reserve} illustrates the scatterplots of $\hat{R}_{k}^{c}$ versus the observed censored reserves given by $R^{c}_k=\sum_{j=t_k+1}^{n}I_{k,j}\times \min \left(Y_{k,j},\ u\right)$. Most of the points are close to the diagonal with a few over or under-estimation for large claims.

\subsection{Results for individual reserving model with large claims }\label{subsec:large_claim_res}

In the following, we adjust the LSTM outputs, presented in Section~\ref{subsec:resultat_real},  using the procedure described in Section~\ref{large_claims} which considers large claims. For the training dataset, Table~\ref{tab:summary_32000} presents the mean, the standard deviation and some quantiles of the LSTM predictions associated to large observed payments censored at \$32,000. The values in Table~\ref{tab:summary_32000} are much smaller than the target $u=32,000$. Indeed, the model prediction is like an expectation of observed incremental payments. Since there are few censored observations, the network tends to understate them. If we had not censored the payments, the discrepancies would have been even greater.
 
\begin{table}[t]
	\centering
	\caption{Summary statistics of LSTM predictions for censored observed large payments.}\label{tab:summary_32000}
	\begin{tabular}{lrrrrr} \hline
		& mean  & std \hspace{0.05cm}  &  $q_{0.25}$ & $q_{0.50}$  &  $q_{0.75}$  \hspace{0.001cm} \\ \hline
		Predicted payment $\hat{Y}$  & 3220 & 4126 &  1002 & 1510 & 2761 
		\\ 
		Expected payment $\hat{p} \times\hat{Y}$ & 2682 & 4216 &  334 & 842 & 2479 \\ \hline
	\end{tabular}
\end{table}

From Table~\ref{tab:summary_32000}, we can see the order of magnitude of the threshold $\zeta$. We propose to apply Eq.~\eqref{formula:LSTM_pred} according to different thresholds $\zeta$, and to compare the adjusted predictions with the uncensored observed payments. Note that in the GLM prediction of the probability of exceedance for future periods, the number of previous periods without payments is approximated as follows when it is unobserved. If $\hat{p}_{k,j}\times \hat{Y}_{k,j}$ is close to the mean of LSTM predicted amounts associated with the actual observed zero in the training dataset, then the period $j$ is considered without payment. For the training, validation and testing datasets, Tables~\ref{tab:ratio_reserve_large} and \ref{tab:ratio_ultimate_large} present the ratio of the aggregate reserve and ultimate, respectively, for several $\zeta$ values. Both are computed from the adjusted predictions and the uncensored observed payments as follows:
\begin{equation*}
 \frac{\sum_{k\in\mathcal{D}}\sum_{j=t_k+1}^{n} \hat{E}[Y_{k,j}]}{\sum_{k\in\mathcal{D}}\sum_{j=t_k+1}^{n} I_{k,j} \times Y_{k,j} } \qquad \text{ and } \qquad \frac{\sum_{k\in\mathcal{D}} \left(\sum_{j=1}^{t_k} I_{k,j} \times Y_{k,j}+ \sum_{j=t_k+1}^{n} \hat{E}[Y_{k,j}]\right)}{\sum_{k\in\mathcal{D}} \sum_{j=1}^{n}I_{k,j} \times Y_{k,j}}.
\end{equation*}
When $ \zeta = 0 $, Eq.~\eqref{formula:LSTM_pred} amounts to adjusting the network outputs to consider large payments occurence for all positive predictions $\hat{Y}_{k,j} $. Unfortunately, this approach grossly overestimates payments, especially for small ones. For $\zeta = \infty $, no adjustment is applied and the network predictions are compared to the actual uncensored payments. Obviously, the ratios are very small since our predictions underestimate large payments that the network has never seen during its training. In addition, we set three values of $\zeta $ that are reasonable based on Table~\ref{tab:summary_32000}. For $\zeta=2000$, the ratios are closer to one compared to $\zeta = 0$, but we still overestimate. Applying Eq.~\eqref{formula:LSTM_pred} with $\zeta=3000$, we underestimate the aggregate reserve by 10\%  for the three datasets. We obtain the best results for threshold $\zeta=2500$, where the ratios of the aggregate reserve and ultimate are very close to one.

Moreover, we compute the ratios using the chain-ladder method, with development factors evaluated from the uncensored training dataset. According to Table~\ref{tab:ratio_reserve_large}, the chain-ladder overestimates the aggregate reserve by 30\%. Our LSTM model with adjustment for large payments using threshold $\zeta=2500$ outperforms the chain-ladder on the aggregate results and has the advantage of yielding individual reserves.

\begin{table}[h]
	\centering
	\caption{Ratios of the aggregate predicted reserve to observed aggregate reserve.}\label{tab:ratio_reserve_large}
	\begin{tabular}{lcccccc}\hline
			Datasets & Chain-ladder & $\zeta=0$ & $\zeta=2000$ & $\zeta=2500$ & $\zeta=3000$ &  $\zeta=\infty$ \\ \hline
		Training  & 1.3479 &  5.5942 &  1.4110 &  \textbf{1.0309} & 0.9048 &  0.7976 \\
		Validation  & 1.3349 & 5.6437 & 1.3623 &   \textbf{0.9911} &0.8808 &0.7850\\ 
		Testing  & 1.3176 & 5.3250 & 1.4426  &  \textbf{1.0547}  &  0.9295 
		& 0.7878 \\ \hline
	\end{tabular} 
\end{table}

\begin{table}[h]
	\centering
	\caption{Ratios of the aggregate predicted ultimate to observed aggregate ultimate. }\label{tab:ratio_ultimate_large}
	\begin{tabular}{lcccccc}\hline
		Datasets & Chain-ladder & $\zeta=0$ & $\zeta=2000$ & $\zeta=2500$ & $\zeta=3000$ &  $\zeta=\infty$ \\ \hline
		Training  & 1.1264 & 2.9361  &  1.1732 & \textbf{1.0130} & 0.9599 & 0.9147 \\
		Validation  & 1.1227 & 3.0057 & 1.1564 &  \textbf{0.9961} & 0.9485 &   0.9071\\ 
		Testing & 1.1193  & 2.8732  & 1.1917  &  \textbf{1.0237} & 0.9694
		& 0.9081 \\ \hline
	\end{tabular} 
\end{table}

\section{Conclusion}\label{sec:conclusion}

In this paper, we propose an individual reserving model based on an LSTM network and a large claim forecasting procedure. Our model has the advantage of being easily adaptable to the structure of any claim dataset with detailed development and of flexibly incorporating large claims. Our procedure involves only the information available at the current evaluation date. Our network shows accurate predictions and outperforms the chain-ladder model in the two case studies carried out with a simulated and a real dataset. 

In forthcoming research work, we are interested in adding other tasks to the network while ensuring a balanced optimization of the loss functions, and in considering model uncertainty. Also, we could further investigate the hyper-parameter selection procedure to enhance network performance, study the uncertainty of predictions and enhance the procedure for handling large claims.

\section*{Acknowledgments}

The authors sincerely thank Christopher Blier-Wong, Julien Fagnan, Luc Lamontagne, Fran\c cois Laviolette, \'{E}tienne Marceau and Fr\'{e}d\'{e}rik Paradis for useful discussions. The authors gratefully acknowledge financial support from the Natural Sciences and Engineering Research Council of Canada (CRDPJ 515901-17, Cossette: RGPIN-2017-04273, C\^{o}t\'{e}: RGPIN-2019-04190) and the Big Data Research Center at Universit\'{e} Laval. We also thank an anonymous insurance company for providing the data. 

\section*{Conflict of interest}

 The authors declare none.

\appendix
\renewcommand{\thesection}{\Alph{section}.\arabic{section}}
\setcounter{section}{0}
\begin{appendices}

\section{Simulated data setup}\label{subsec:appA}

\renewcommand{\thefigure}{A.\arabic{figure}}
\renewcommand{\thetable}{A.\arabic{table}}
\setcounter{figure}{0}
\setcounter{table}{0}

The \cite{gabrielli2018individual} simulator must be provided with a few parameters such as the expected number of claims, a probability distribution for the allocation of claims to the lines of business (LoB), growth rates for the number of claims over the years for each LoB, and standard deviations for the total individual claim amounts and recoveries. Table~\ref{tab:para_generator} presents the values of these parameters in the case study from Section~\ref{sec_simulated}.

\begin{table}[h]
	\centering
	\caption{Parameters 
		used to generate the simulated dataset.}\label{tab:para_generator}
	\begin{tabular}{lc}\hline
		\hspace{2cm} Parameter & Value \\ \hline
		Expected number of claims & 1,000,000 \\ 
		Probability distribution for claim allocation  & $\Pr(\text{LoB}=1)=\Pr(\text{LoB}=4)=0.25, $  \\ 
		& $\Pr(\text{LoB}=2)=0.30$,  $\Pr(\text{LoB}=3)=0.20 $\\
		Growth rate for the number of claims  & 0.01 (same for all lines of business)\\
		Standard deviation   & 0.85 (for payments and recoveries)\\ \hline
	\end{tabular}
\end{table}

\section{Description of the real dataset}\label{subsec:appB}

\renewcommand{\thefigure}{B.\arabic{figure}}
\renewcommand{\thetable}{B.\arabic{table}}
\setcounter{figure}{0}
\setcounter{table}{0}

\begin{table}[H]
	\centering
	\caption{List of static covariates in the real dataset used in our model.} \label{tab:list_static_real}
	\begin{tabular}{lcccc}\hline
		\hspace{0.6cm} Static features & Categorical & $\#$ of categories & Missing values & Pre-processing \\ \hline
		Day of occurrence  & \xmark & --- &  \xmark  & ---\\
		Month of occurrence  & \xmark &  --- &  \xmark  & --- \\
		Year of occurrence & \xmark & --- & \xmark & Shifted with 2006  \\
		Reporting delay  (days)& \xmark & --- & \xmark & --- \\
		Company name &  $\checkmark$ & 5  & \xmark & Indexed \\
		Claimant age &  \xmark & --- & $\checkmark$ & Scaled to $[0, 1]$ \\
		Claimant gender (indicator) &  $\checkmark$ &2 &  $\checkmark$ & --- \\
		Driver experience (years) & \xmark & --- & $\checkmark$ & --- \\
		Vehicle type (indicator) &  $\checkmark$ & 2 & $\checkmark$ &  --- \\
		Vehicle age (years) & \xmark & --- &$\checkmark$ & ---  \\
		Drive train & \xmark & --- & $\checkmark$ & ---  \\
		Number of cylinders & \xmark & --- & $\checkmark$ & --- \\
		Vehicle body & $\checkmark$ & 11 & $\checkmark$ &Indexed  \\
		Convertible (indicator) & $\checkmark$ & 2 & $\checkmark$ & --- \\
		Claim on highway & $\checkmark$ & 4 & \xmark & Indexed \\
		Claim province 	 & $\checkmark$ & 11 & $\checkmark$ & Indexed   \\ \hline
	\end{tabular}
\end{table}

\begin{table}[H]
	\centering
	\caption{List of dynamic covariates in the real dataset used in our model.}\label{tab:list_dynamic_real}
	\begin{tabular}{lc}\hline
		\hspace{0.7cm}  Dynamic features &  Pre-processing  \\ \hline
		Degree of claimant injury &   --- \\
		Incremental monthly payment &  Centered and scaled as in Eq. \eqref{transform} \\
		Number of claimants &  --- \\
		Number of reported injuries  & ---  \\ \hline
	\end{tabular}
\end{table}

\begin{figure}[h]
	\centering
	\begin{subfigure}{0.4\textwidth}
		\includegraphics[scale=0.45]{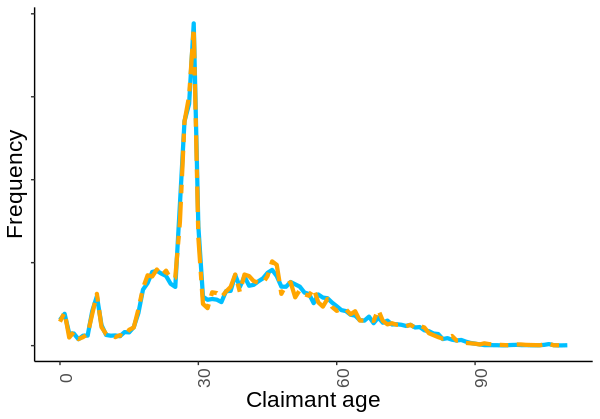}
		\subcaption{}\label{age}
	\end{subfigure}
	\begin{subfigure}{0.3\textwidth}
		\includegraphics[scale=0.45]{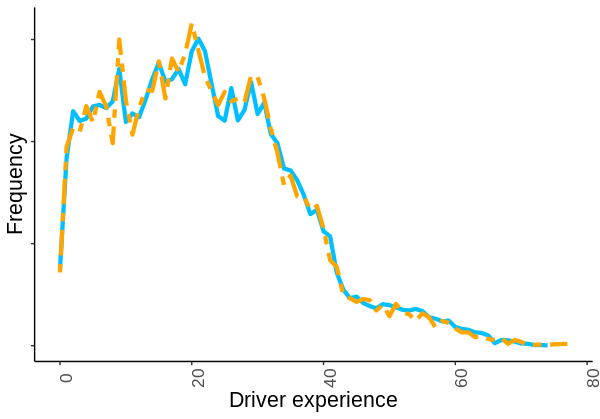}
		\subcaption{}\label{drv}
	\end{subfigure}
	\begin{subfigure}{0.3\textwidth}
		\includegraphics[scale=0.45]{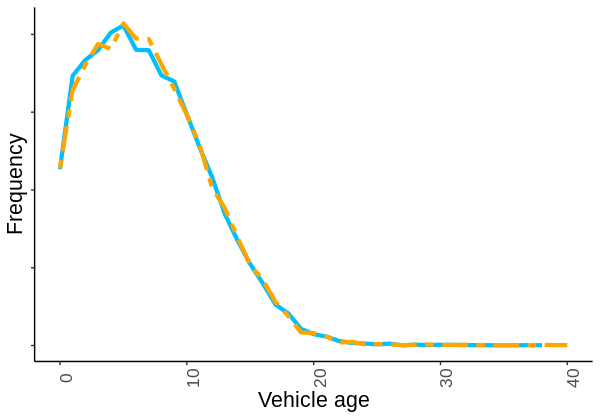}
		\subcaption{}\label{veh}
	\end{subfigure}
	\caption{Frequency of claimant age (a), driver experience (b) and vehicle age (c), for training (solid blue) and validation (dashed orange) datasets. The $y$ axes are masked for confidentiality.}\label{graphe_analyse}
\end{figure}

Figure~\ref{graphe_analyse} presents the empirical frequency of three continuous variables in the real dataset studied in Section~\ref{sec_real_data}: claimant age, driver experience, and vehicle age. We observe similar curves for training and validation datasets. According to Figure~\ref{graphe_analyse}(\subref{age}), young drivers are the riskiest due to their short driving experience. This observation is also supported by Figure~\ref{graphe_analyse}(\subref{drv}), illustrating the decrease in claim occurrence with driving experience. According to Figure~\ref{graphe_analyse}(\subref{veh}), a new vehicle is riskier than an older one. This can be explained by the vehicle's power, encouraging excess speed, or fewer old cars on the road.

\section{Diagnostic plots for extreme value model}\label{subsec:appC}
\renewcommand{\thefigure}{C.\arabic{figure}}
\renewcommand{\thetable}{C.\arabic{table}}
\setcounter{figure}{0}
\setcounter{table}{0}

\begin{figure}[h]
	\centering
	\includegraphics[scale=0.6]{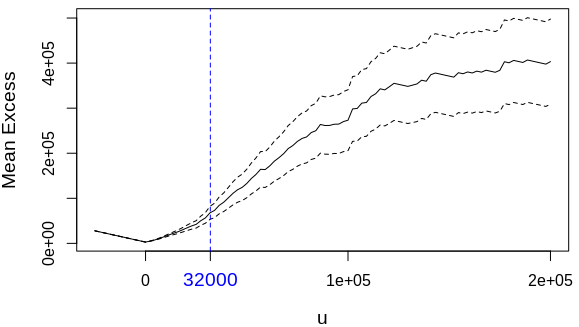}
	\caption{Mean excess plot for incremental payments.}\label{mean_plot}
\end{figure}

\begin{figure}[h]
	\centering
	\includegraphics[scale=0.6]{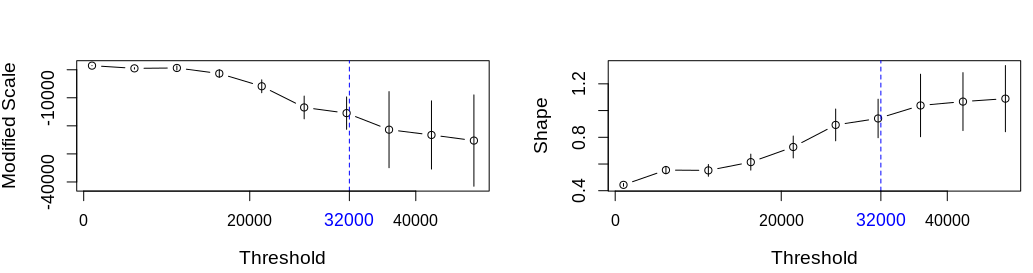}
	\caption{Parameter estimates against threshold for incremental payments.}\label{parameter_stab}
\end{figure}

Figure~\ref{mean_plot} represents the mean excess plot with approximate 95\% confidence intervals for the incremental payments $\boldsymbol{Y}$ as defined in Section~\ref{large_claims}.  The curve is approximately linear between $32,000$ and $60,000$, then it decays sharply.  For extreme values of $u$, the mean excess plot is unreliable due to the limited number of observations on which the estimate and confidence interval are based. Given that the 0.995 quantile of $\boldsymbol{Y}$ is around 31,600, we set $u=32,000$.  This choice makes it possible to keep enough extreme observations for a reliable estimation of the GPD parameters. We use a second tool to confirm this choice, which consists in fitting the GPD at a range of thresholds and looking for parameter estimates stability. Figure~\ref{parameter_stab} depicts the plots of the scale $\sigma$ and shape $\lambda$ estimates against thresholds with 95\% confidence bounds. We select the threshold $u$ as the lowest value for which the estimates remain near-constant. Again, $u = 32,000$ appears appropriate.

\begin{figure}[H]
	\centering
	\includegraphics[scale=0.5]{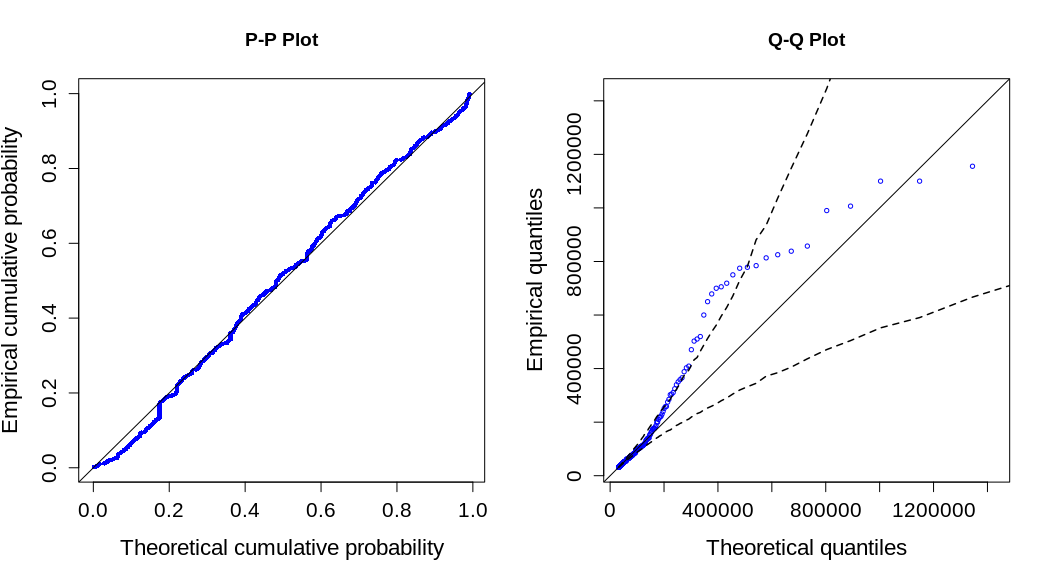}
	\caption{Diagnostic plots for threshold excess model fitted to incremental payments.}\label{fig:pp_qq}
\end{figure}

To assess the goodness of fit of the estimated GPD to $\left(Y_{k,j}-u|Y_{k,j}>u\right)$, we use the cumulative probability (P-P) and quantile (Q-Q) plots. If the estimated GPD model fits the exceedance observations well, then both plots
should be approximately linear. Figure~\ref{fig:pp_qq} shows both diagnostic plots, with points approximately aligned for the P-P plot. In the Q-Q plot, the points form a line for values below $200,000$ ($0.9995$ quantile), and deviate from the diagonal for some large exceedance values outside the 95\% confidence interval. However, we have too few exceedances above $200,000$ to make meaningful conclusions. Thus, both plots validate the estimated GPD model.

\end{appendices}

\bibliographystyle{ama}
\bibliography{references}

\end{document}